\pdfoutput=1

\documentclass[11pt]{article}

\usepackage[]{style/acl}

\usepackage{times}
\usepackage{latexsym}

\usepackage{chngcntr}

\usepackage[T1]{fontenc}

\usepackage[utf8]{inputenc}

\usepackage{microtype}
\usepackage[normalem]{ulem}
\useunder{\uline}{\ul}{}
\title{Coherence boosting: \\ When your pretrained language model is not paying enough attention}

\author{Esmeralda S. Whitammer\\
  Mila / Universit\'e de Montr\'eal \\
  \texttt{\small esmeralda.whitammer@mila.quebec} \\\And
  Zhen Wang \\
  Ohio State University \\
  \texttt{\small wang.9215@osu.edu} \\\And
  Nebojsa Jojic\\
  Microsoft Research \\
  \texttt{\small jojic@microsoft.com}
  }

\usepackage{booktabs}
\usepackage{graphicx}
\usepackage{hyperref}
\usepackage{url}
\usepackage{amsmath,amsfonts,amssymb}
\usepackage{newtxmath}
\usepackage{ulem}
\usepackage{multicol}
\usepackage{enumerate}
\renewcommand{\emph}{\textit}
\usepackage{caption}
\usepackage{subcaption}
\usepackage{xspace}

\usepackage{lipsum,multirow}

\newcommand{\EE}{\mathbb{E}}
\newcommand{\DDD}{\mathcal{D}}
\newcommand{\LLL}{\mathcal{L}}

\DeclareMathOperator{\softmax}{softmax}
\newcommand{\gq}[1]{\left\langle#1\right\rangle}
\newcommand{\bq}[1]{\left[#1\right]}
\newcommand{\pq}[1]{\left(#1\right)}
\newcommand{\aalpha}{\boldsymbol{\alpha}}
\newcommand{\llambda}{\boldsymbol{\lambda}}
\newcommand{\bbeta}{\boldsymbol{\beta}}
\renewcommand{\AA}{\mathbf{A}}
\newcommand{\nop}[1]{}

\begin{document}

\maketitle

\begin{abstract}
Long-range semantic coherence remains a challenge in automatic language generation and understanding. We demonstrate that large language models have insufficiently learned the effect of distant words on next-token prediction. We present \emph{coherence boosting}, an inference procedure that increases a LM's focus on a long context. We show the benefits of coherence boosting with pretrained models by distributional analyses of generated ordinary text and dialog responses.
It is also found that coherence boosting with state-of-the-art models for various zero-shot NLP tasks yields performance gains with no additional training.
\end{abstract}

\section{Introduction}

\renewcommand{\thefootnote}{\relax}

Language models (LMs) are commonly evaluated for their ability to generate, rank, or classify coherent spans of text. 
\textbf{Long-range semantic coherence} is a unifying feature of modern NLP benchmarks and applications, whether they are about producing short answers to questions, ranking answer choices by their consistency with world knowledge, or generating long responses.\footnote[0]{Code: \href{https://github.com/zhenwang9102/coherence-boosting}{github.com/zhenwang9102/coherence-boosting}.}

\renewcommand{\thefootnote}{\arabic{footnote}}


Large nonspecialized LMs, such as GPT-2 and -3 \citep{radford2019language,brown2020language}, sometimes fail to understand or use the semantic link between a text and its prompt or long-range context  (Fig.~\ref{fig:bifocal_distributions}).
Samples from these LMs have an unnaturally low density of words that require many tokens of context to predict (\S\ref{sec:experiments_owt}), and the scores that the models give to completions of prompts indicate that they are oversensitive to recent context (\S\ref{sec:experiments_nlu}).

We hypothesize that these failures arise from  modeling choices and distribution shift. 
Specifically, autoregressive LMs are typically fit to a multi-objective problem: simultaneously maximizing token likelihoods conditioned on many lengths of truncated context (\S\ref{sec:why_cb}). 
Yet, at generation or scoring time, likelihoods are conditioned on the entire prompt or previously generated string, specifically selected
to be coherent or even guaranteed to influence the output.
The two common solutions -- finetuning models on one or multiple tasks \citep{khashabi-etal-2020-unifiedqa,sanh-etal-2022-multitask} and improving models or prompts to facilitate in-context learning \citep{brown2020language,schick-schutze-2021-exploiting} -- do not directly target the problem of long-range coherence.

This paper proposes \textbf{coherence boosting}, a simple inference-time procedure that increases the effect of distant words on predicted token distributions and is applicable in both generation and ranking settings. A pretrained model is viewed as an \emph{ensemble} of experts that produce token distributions conditioned on varying lengths of context. These experts are log-linearly mixed to form a predictor that is superior to the base model (\S\ref{sec:cb}).

Coherence boosting greatly improves prediction of words that depend on a long context, as evidenced by state-of-the-art results on tasks specially meant to assess models' attention to distant words (\S\ref{sec:lambada}).
In generation of generic text and dialog responses, we show that coherence boosting brings the frequency of occurrence of such words close to that seen in natural text (\S\ref{sec:experiments_generation}). Beyond generation, we study diverse multiple-choice tasks (\S\ref{sec:experiments_nlu}), in which examples are known to be highly coherent. Coherence boosting does not modify the base model and depends on a single parameter than can be estimated in one pass through a validation set, yet is a competitive adaptation algorithm.

\begin{figure*}[t]
\newcommand\blueuline{\bgroup\markoverwith
{\textcolor{blue}{\rule[-0.75ex]{2.8pt}{0.7pt}}}\ULon}

\newcommand\reduline{\bgroup\markoverwith
{\textcolor{red}{\rule[-1.2ex]{0.1pt}{0.7pt}}}\ULon}

\newcommand\phuline{\bgroup\markoverwith
{\textcolor{white}{\rule[-1.2ex]{0.1pt}{0.7pt}}}\ULon}

\def\word{\textcolor{gray}{\rm w}}
\def\full{\textcolor{blue}{\rm full}}
\def\short{\textcolor{red}{\rm short}}
\newcommand{\mybox}[1]{#1}
\definecolor{f}{rgb}{0.97,0.97,1.0}

\hspace{-8pt}
\colorbox{f}{
\begin{minipage}{1\linewidth}
\resizebox{0.9\linewidth}{!}{
\begin{minipage}{1\linewidth}

\textbf{A: }\phuline{\blueuline{I'm Natasha. I study neural language models and dialog systems. Are you an AI researcher too?}} \\
\textbf{B: }\reduline{\blueuline{No, though I do like chatting with bots and laughing at their mistakes. But what was your name again?}}\\
\textbf{A: }\reduline{\blueuline{Oh, you forgot already? My name is}} \fbox{$\word$} \\[-4pt]

\begin{tabular}{rl}
$p_{\full}=f(\word\mid\full)$ 
&
\textit{1.} Alex (1.9\%)
\textit{2.} \textbf{Natasha} (1.7\%)
\textit{3.} also (1.5\%)
\\
$p_{\short}=f(\word\mid\short)$ 
&
\textit{1.} : (3.4\%)
\textit{2.} the (1.9\%)
\textit{3.} in (1.2\%)
 \hfill\dots
\textit{3358.} \textbf{Natasha} (0.0042\%)
\\
$p_{\full}^{1.5}p_{\short}^{-0.5}$
& 
\multicolumn{1}{l}{
\textit{1.} \textbf{Natasha} (20.5\%)
\textit{2.} Alex (2.2\%)
\textit{3.} Nat (2.1\%)
}
\end{tabular}\\[10pt]

\centering
\blueuline{Ballad metre is ``less}\reduline{\blueuline{ regular and more conversational'' than common}} \fbox{$\word$} \\[2pt]

\begin{tabular}{rl}
$p_{\full}=f(\word\mid\full)$ 
&
\textit{1.} sense (9.0\%)
\textit{2.} in (2.0\%)
\textit{3.} . (1.9\%)
 \hfill\dots
\textit{13.} \textbf{metre} (0.6\%)
\\
$p_{\short}=f(\word\mid\short)$ 
&
\textit{1.} sense (7.8\%)
\textit{2.}  English (3.5\%)
\textit{3.} . (3.2\%)
 \hfill\dots
\textit{14103.} \textbf{metre} (0.00014\%)
\\
$p_{\full}^{1.5}p_{\short}^{-0.5}$
& 
\multicolumn{1}{l}{
\textit{1.} \textbf{metre} (16.2\%)
\textit{2.} sense (4.0\%)
\textit{3.} meter (2.5\%)
}
\end{tabular}\\[10pt]

\blueuline{Isley Brewing Company: Going}\reduline{\blueuline{ Mintal -- a minty milk chocolate}} \fbox{$\word$} \\[2pt]

\begin{tabular}{rl}
$p_{\full}=f(\word\mid\full)$ 
&
\textit{1.} bar (4.8\%)
\textit{2.} drink (3.7\%)
\textit{3.} with (3.5\%)
 \hfill\dots
\textit{13.} \textbf{stout} (2.7\%)
\\
$p_{\short}=f(\word\mid\short)$ 
&
\textit{1.} bar (6.9\%)
\textit{2.}  that (5.7\%)
\textit{3.} , (4.4\%)
 \hfill\dots
\textit{60.} \textbf{stout} (0.23\%)
\\
$p_{\full}^{1.5}p_{\short}^{-0.5}$
& 
\multicolumn{1}{l}{
\textit{1.} \textbf{stout} (7.4\%)
\textit{2.} ale (5.6\%)
\textit{3.} bar (3.1\%)
}
\end{tabular}\\[10pt]

\blueuline{Other times anxiety is not as easy to}\reduline{\blueuline{ see, but can still be just as}} \fbox{$\word$} \\[2pt]

\begin{tabular}{rl}
$p_{\full}=f(\word\mid\full)$ 
&
\textit{1.} important (5.6\%)
\textit{2.} bad (4.6\%)
\textit{3.} \textbf{debilitating} (4.3\%)
\\
$p_{\short}=f(\word\mid\short)$ 
&
\textit{1.} effective (16.2\%)
\textit{2.}  good (7.4\%)
\textit{3.} useful (3.9\%)
 \hfill\dots
\textit{294.} \textbf{debilitating} (0.035\%)
\\
$p_{\full}^{1.5}p_{\short}^{-0.5}$
& 
\multicolumn{1}{l}{
\textit{1.} \textbf{debilitating} (17.6\%)
\textit{2.} real (6.0\%)
\textit{3.} severe (5.8\%)
}
\end{tabular}

\end{minipage}
}
\end{minipage}
}
\caption{Next-token probabilities given by LMs (DialoGPT and GPT-2) conditioned on a \textcolor{blue}{long context} and on a \textcolor{red}{partial context}. The top words in both distributions are incorrect, but a log-linear mixture of the distributions makes the correct word most likely. Sampling from such a mixture at each generation step (\emph{coherence boosting}) improves the quality of output text (\S\ref{sec:experiments_generation}). (Dialog example constructed by the authors; other examples from OpenWebText.)}
\label{fig:bifocal_distributions}
\end{figure*}

\subsection{Background and related work}

Balance between satisfaction of short-range statistical constraints and maintenance of long-range structure was a central question of language generation long before neural language modeling. To compensate for the sparsity of the learning signal for long-range influences, $n$-gram models and early neural language models used `backing-off' schemes that interpolate between predictors with different context lengths \citep{chen-goodman-1996-empirical,bengio2003neural}.
Neural language modeling brought a need for recurrent units with better numerical properties for propagating information over long distances  \citep{hochreiter1997long,cho-etal-2014-learning} and eventually saw the reintroduction of alignment variables \citep{brown-etal-1993-mathematics} into generation in the form of attention \citep{bahdanau2015neural,vaswani2017attention}. Attention is at the core of Transformer LMs, including GPT.

Language models are being trained on and adapted to ever-longer input sequences \citep{beltagy2020longformer, zaheer2020big, roy-etal-2021-efficient,press2021train}, but they remain undersensitive to distant content or syntax \citep{khandelwal-etal-2018-sharp,sun-etal-2021-long} and are easily fooled by recency bias in few-shot prompts \citep{zhao2021calibrate} or multi-turn conversations \citep{sankar-etal-2019-neural}. 

Recent work has continued to study inference-time procedures that prevent text sampled from LMs from degenerating into nonsense. Most of these procedures, such as tempered sampling and top-$k$/top-$p$ truncation \citep{fan-etal-2018-hierarchical, holtzman2020curious}, independently modify the output distribution at each generation step to decrease its entropy and diminish its low-likelihood tail. \citet{holtzman2020curious} and  \citet{meister-cotterell-2021-language} found that such local modifications increase the quality of long generated sequences; we adopt and extend their methodology in \S\ref{sec:experiments_owt}. 

For dialog systems, \citet{li-etal-2016-diversity} propose a decoding scheme that maximizes a mutual information criterion, which explicitly optimizes for dependence of generated text on prompts -- a special case of coherence boosting.
In multiple-choice tasks, where a model must choose one of several given completions of a prompt, \citet{brown2020language} observe that selecting the completion that maximizes the conditional likelihood of the completion following the prompt often favors completions having high \emph{unconditional} likelihood (likelihood following an empty or dummy prompt) and, for some tasks, chooses to divide the scores of candidate answers by their unconditional likelihoods. This 
is also a special case of coherence boosting.

Such scoring modifications are more thoroughly studied by \citet{zhao2021calibrate,holtzman2021surface}. The latter attributes the problem to `surface form competition': there are many variants of the correct completion that together may capture a large part of probability mass, but the form of the given answer choice alone is not the most likely. However, we show that other causes are at play: surface form competition is impossible when the completion is known to be a single token and the range of choices is the whole vocabulary (\S\ref{sec:lambada}), and it is not applicable to open-ended generation (\S\ref{sec:experiments_generation}).

\section{Coherence boosting}
\label{sec:cb}

In this section, $f$ is an autoregressive LM over a vocabulary $V$ with learnable parameters $\theta$, taking as input a variable number of tokens (up to a maximum context length $M$) and producing a vector of next-token likelihoods:
\[
f(w_1,\dots,w_n;\theta)\in\Delta(V),\quad w_1,\dots,w_n\in V,
\]
where $\Delta(V)$ is the probability simplex over $V$. 
We will write the $w$-th component of this output vector as a conditional likelihood, $f(w\mid w_1,\dots,w_n;\theta)$.

We denote by $f_k$ the model evaluated on only the \emph{last} $k$ input tokens, ignoring earlier tokens:
\[
f_k(w_1,\dots,w_n;\theta):=f(w_{n-k+1},\dots,w_n;\theta).
\]
\paragraph{Coherence boosting for next-token prediction.} \emph{Coherence boosting} for a model $f$ selects real-valued weights $\aalpha=(\alpha_1,\alpha_2,\dots,\alpha_M)$ and produces a new language model $f_{\aalpha}$, defined by
\begin{align}
    &f_{\aalpha}(w_1,\dots,w_n;\theta)\nonumber\\&:=\softmax\pq{\sum_{k=1}^M\alpha_k\log f_k(w_1,\dots,w_n;\theta)},\label{eqn:boosting_gen}
\end{align}
where $\log$ is taken element-wise, or, equivalently,
\[
    f_{\aalpha}(w|w_1,\dots,w_n;\theta)\propto\prod_{k=1}^Mf_k(w|w_1,\dots,w_n;\theta)^{\alpha_k}.
\]
This is a weighted product-of-experts model, where the `experts' are copies of the base model $f$ evaluated on different context lengths. 

Because evaluating $f$ is expensive, we use sparse weights $\aalpha$, as the expression (\ref{eqn:boosting_gen}) depends only on those $f_k$ for which $\alpha_k\neq0$. In Fig.~\ref{fig:bifocal_distributions} and in the experiments, we allow $\aalpha$ to have only two nonzero entries: when computing likelihoods of words following a sequence of length $n$, we consider weighted products of $f_{\rm max}:=f_n$ (the full context) and an $f_k$ with $k\leq n$ (a short context, either of fixed length or decided by prompt structure as in \S\ref{sec:experiments_dialog}).

As its name suggests, the form of coherence boosting in (\ref{eqn:boosting_gen}) bears a resemblance to log-linear boosting for multiclass classification \citep{friedman2000additive}. However, our weak classifiers are pretrained and share all of their parameters, not obtained by an iterative procedure of training on reweighted data, and we permit negative weights.\footnote{As for the first half of the term `coherence boosting', \citet{howcroft-etal-2020-twenty,belz-etal-2020-disentangling} found that very incoherent definitions of the word `coherence' abound in the natural language evaluation literature. The reader is asked to forgive us for the loose definition of `long-range semantic coherence' adopted in this paper.}

\paragraph{Coherence boosting for answer selection.} 

In multiple-choice problems, a LM must choose the best answer following a context, which consists of a premise or passage followed by a shorter \emph{premise-free context} (either a short phrase, such as ``Answer:'', that incites the LM to generate an answer in the right format, or a hypothesis that depends on the premise). The full context is the concatenation of the premise and the premise-free context (\S\ref{sec:prompt_formats}).

By the autoregressive factorization, the model $f$ assigns conditional likelihoods to \emph{sequences} of tokens following context. A typical model for answer selection ranks the candidate answers $a_i$ (sequences of tokens) by
$f(a_i\mid \textrm{full context};\theta)$ and outputs the highest-ranked $a_i$. \emph{Coherence boosting} chooses a parameter $\alpha$ and ranks the choices by:
\begin{align}
    &\log f(a_i\mid\textrm{full context};\theta)\,+\nonumber\\&+ \alpha\log f(a_i\mid\text{premise-free context};\theta).
    \label{eqn:boosting_mc}
\end{align}
This is a log-linear combination of two models: $f$ evaluated with full context and with a partial context. When $\alpha=0$, ranking by (\ref{eqn:boosting_mc}) is equivalent to ranking by the base model. When $\alpha=-1$, it is equivalent to dividing the base model's score by the score of each answer conditioned on the prompt (short context), and thus to maximizing pointwise mutual information between the premise and the answer conditional on the premise-free context. Unlike \citet{brown2020language,holtzman2021surface}, our formulation allows the premise-free context to include information specific to the example, not only a domain-specific dummy prompt.

We expect coherence boosting to correct for an oversensitivity to the premise-free context, and thus the optimal $\alpha$ will typically be negative (see \S\ref{sec:experiments_nlu}).

\subsection{Why should boosting models be better than full-length predictors?}
\label{sec:why_cb}

\paragraph{Multi-objective training.} 

As we will now see, the training of the model $f$ simultaneously fits all of the predictors $f_k$, which share parameters $\theta$. Each training iteration samples a sequence (or batch of sequences) of a chosen maximum length $M+1$ from the data distribution $\DDD$ and minimizes the average negative log-likelihood (NLL) of \emph{all} words following the parts of the sequence that precede them: the optimization criterion is:
\[
\EE_{w_1\dots w_{M+1}\sim\DDD}{\frac1M\sum_{k=1}^M-\log f(w_{k+1}|w_1,\dots,w_k;\theta)}.
\]
If $\DDD$ is uniform over all length-($M+1$) subsequences of a training corpus, any given word is equally to likely to appear in all positions within a sampled sequence\footnote{Many authors leave unspecified the way in which training batches are formed from a corpus of input documents. Here we assume that all training documents are concatenated into one (very long) document separated by end-of-text tokens and ignore minute effects near the start and end of this document.}, and the criterion is equal to
\begin{align}
&\sum_{k=1}^M\frac1M\underbrace{{\EE\left[-\log f_k(w_{M+1}|w_1,\dots,w_M;\theta)\right]}}_{\LLL_k(\theta)},\label{eqn:criterion_scalarization}
\end{align}
This is a uniform scalarization of an $M$-task problem: the $k$-th objective $\LLL_k(\theta)$ is the expected NLL of a word in the corpus following $k$ context words.

This situation is different from that seen at \textit{generation} time. If the text generated so far is $w_1w_2\dots w_n$, the distribution from which the next word $w_{n+1}$ is sampled is $f_n(w_1,\dots,w_n;\theta)$ -- only the ensemble member using full context is used. However, if the string $w_1\dots w_nw_{n+1}$ had been seen in training, $f$ would have been trained to predict $w_{n+1}$ given \textit{all partial contexts}, with equal weight given to all prediction losses. 
Thus, $f$ is trained to make predictions on data it never sees in evaluation, and may be prevented from optimally learning to use long context: 
parameters that locally optimize (\ref{eqn:criterion_scalarization}) are locally Pareto-optimal for the \emph{set} of prediction losses $\LLL_1,\dots,\LLL_M$, but not necessarily optimal for any individual $\LLL_k$. An ensemble of the $f_k$ ($k\leq n$) may be a better predictor than $f_n$ alone. (See \S\ref{sec:derivation} for further analysis of when this occurs.)

\paragraph{Undertraining.} The parameters $\theta$ are shared by the predictors $f_k$, and modeling power must be spread among the losses $\LLL_k(\theta)$. The short-context predictors are easier to fit, while sequences in which long context affects the prediction are rare. We expect sensitivity to long context, and precision in modeling its effect, to be especially diminished if the model is undertrained.

\paragraph{Distribution shift.} 
While the training procedure causes a bias against the influence of longer contexts on generation, we see the opposite bias in downstream tasks (question answering, natural language inference, adversarial probes for common sense): 
Many modern NLP benchmarks try to challenge models to use long context (\S\ref{sec:lambada}, \S\ref{sec:experiments_nlu}).

\section{Experiments: LAMBADA}
\label{sec:lambada}

\begin{table*}[t!]
    \vspace*{-1em}
\centering
\resizebox{0.7\linewidth}{!}{

\begin{tabular}{lcccccccc}
\toprule
& \multicolumn{4}{c}{GPT-2} & \multicolumn{4}{c}{GPT-3} \\\cmidrule(lr){2-5}\cmidrule(lr){6-9}
           & 125M  & 350M  & 760M  & 1.6B  & 2.7B  & 6.7B  & 13B   & 175B  \\
           \midrule
$f_{\max}$ & 47.66 & 57.29 & 61.23 & 64.25 & 62.39 & 71.40 & 76.58 & 81.51 \\
CB ($\alpha_k=\alpha_k^*$)   & 66.70 & 73.53 & 76.54 & 77.53 & 77.00 & 81.84 & 86.36 & 88.61 \\ 
\midrule
$\alpha_k^*$  & $-$0.6   & $-$0.5   & $-$0.5   & $-$0.5   & $-$0.3   & $-$0.3   & $-$0.3   & $-$0.2   \\
$k^*$       & 10    & 11    & 10    & 9     & 9     & 10    & 3     & 3    \\
\bottomrule
\end{tabular}

}
\caption{Accuracy (\%) and optimal boosting parameters on LAMBADA: $f_{\rm max}$ is the full-context model without boosting; CB is our model with the optimal boosting parameters (last two rows).}
\label{tab:lambada}
\end{table*}

\begin{figure}[t!]
\centering
\includegraphics[width=0.4\textwidth]{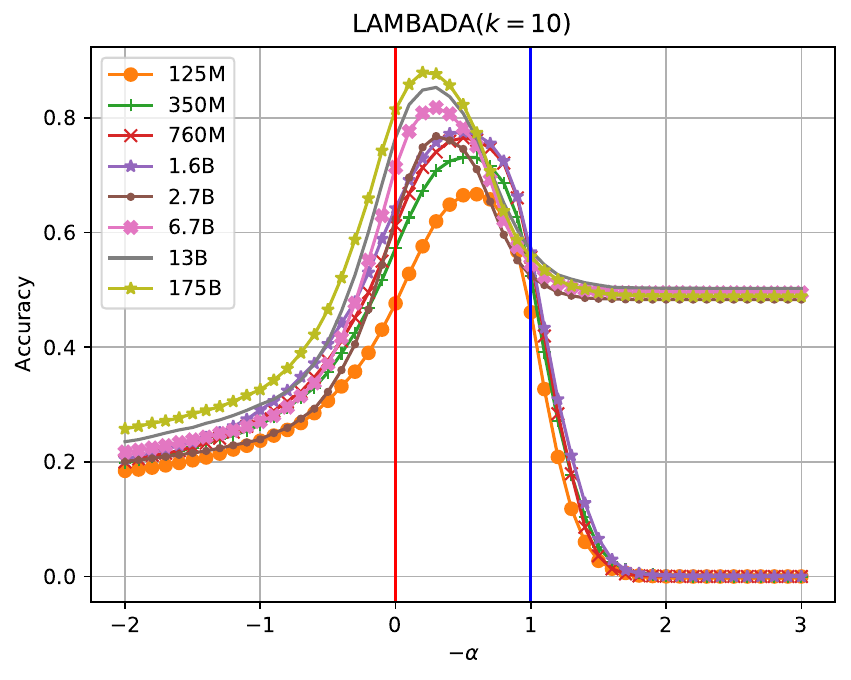}
\vspace*{-1em}
\caption{Model comparison on LAMBADA with $k=10$ and varying $\alpha_k$. The red line ($\alpha=0$) is the base LM $f_{\rm max}$.  (The different right tails of GPT-3 models are due to top-100 truncation of logits returned by the API.)}
\label{fig:lambada_alpha}
\end{figure}

The LAMBADA dataset~\citep{paperno-etal-2016-lambada} tests LMs' understanding of long-range dependencies by measuring the prediction of the final words in passages of several sentences. The task explicitly requires reasoning over a broad context: humans can reliably guess the last word when given a whole passage, but not when given only the last sentence. 

We perform experiments with the GPT family of models, closely replicating the evaluation setting of \citet{radford2019language}.\footnote{Certain details are omitted by \citet{radford2019language}. Based on \url{https://github.com/openai/gpt-2/issues/131}, we nearly match baseline accuracy by predicting the last subword token, rather than the last word.} 
We predict the final word as the top-ranked token under the boosted model $f_{\rm max}f_k^{\alpha_k}$, where $f_{\rm max}$ is the model taking the full available context and $k,\alpha_k$ are the chosen length and coefficient of the short context. To choose $k$ and $\alpha_k$, we do a grid search on the validation set and apply the best values to the testing set.

\paragraph{Results.}
Table~\ref{tab:lambada} shows the accuracies and optimal parameter values $k^*,\alpha_k^*$. Coherence boosting vastly reduces prediction error for all models. In particular, the boosted GPT-2 Small performs better than the original GPT-3 2.7B. 
The boosted GPT-3 175B achieves a new state of the art. 

Other than the impressive performance gain, we highlight two observations. \textbf{(1)} The optimal $\alpha_k$ is always negative, indicating that the optimal mixture of models penalizes the influence of short-range context relative to long-range context. \textbf{(2)} With increasing model size, the optimal $\alpha_k$ and $k$ become closer to 0. This means that bigger models capture long-range coherence better than small models, as they have less need to penalize the effect of short context.  (Fig.~\ref{fig:lambada_alpha} shows the accuracy curves for all models by sweeping $\alpha_k$ with a fixed $k$. The peak clearly moves to the left as model size grows.)

\section{Experiments: Language generation}
\label{sec:experiments_generation}

\subsection{Generic text}
\label{sec:experiments_owt}

\begin{table*}[t]
\centering
\scalebox{0.9}{
\begin{tabular}{lcccccccc}
\toprule
& \multicolumn{4}{c}{from \citet{holtzman2020curious}} & \multicolumn{2}{c}{lex coherence} & \multicolumn{2}{c}{long-dep tokens} \\ \cmidrule(lr){2-5}\cmidrule(lr){6-7}\cmidrule(lr){8-9}
    Inference method & ppl & BLEU-4 & Zipf & rep \% & LR$_{50}$ \% & LR$_{100}$ \% & $\delta$ \% & LTF \% \\
    \midrule
    Sampling & 23.53 & 0.28 & {\textbf{0.93}} & {\textbf{0.22}} & 12.92 & 7.71 & 4.87 & 3.28  \\
    Sampling ($T=0.9$) & 10.60 & 0.35 &  0.96 & 0.66 & {16.36} & 10.01 & {\textbf{6.54}} & 4.15 \\
    Nucleus ($p=0.95$) & 13.48 & 0.32  & 0.95 & 0.46 & 15.06 & 9.11 & 5.65 & 3.62  \\ \midrule
    + boost ($k=32$, $\alpha_k=-0.05$) & 12.81 & \textit{\textbf{0.31}} & \textit{{0.94}} & \textit{\textbf{0.34}} & \textit{15.54} & \textbf{\textit{9.42}} & \textit{6.16} & \textbf{\textit{3.98}} \\
    + boost ($k=64$, $\alpha_k=-0.1$) & \textit{\textbf{12.93}} & \textit{0.32} & \textit{0.95 }& \textit{0.46} & \textbf{\textit{15.75}} & \textit{9.67} & \textit{6.10} & \textit{3.95} \\ \midrule
    + self-tune (\S\ref{sec:coherence_tuning})
    & 10.16 & 0.33 & 0.95 & 0.64 & 16.19 & 9.85 & 6.59 & 4.16 \\ \midrule
    Human & 13.19 & 0.31 & 0.93 & 0.28  & 15.95 & 9.51 & 6.54 & 4.03  \\\bottomrule
\end{tabular}
}

\caption{Distributional metrics of WebText completions. The last four columns are measures of long-range coherence (\S\ref{sec:experiments_owt}). (Nearest-to-human values in \textbf{bold}, boosting models better than top-$p$ sampling alone in \textit{italics}.)}
\label{tab:owt-gen}
\end{table*}

\begin{figure*}[t]
\centering
\includegraphics[width=0.95\textwidth]{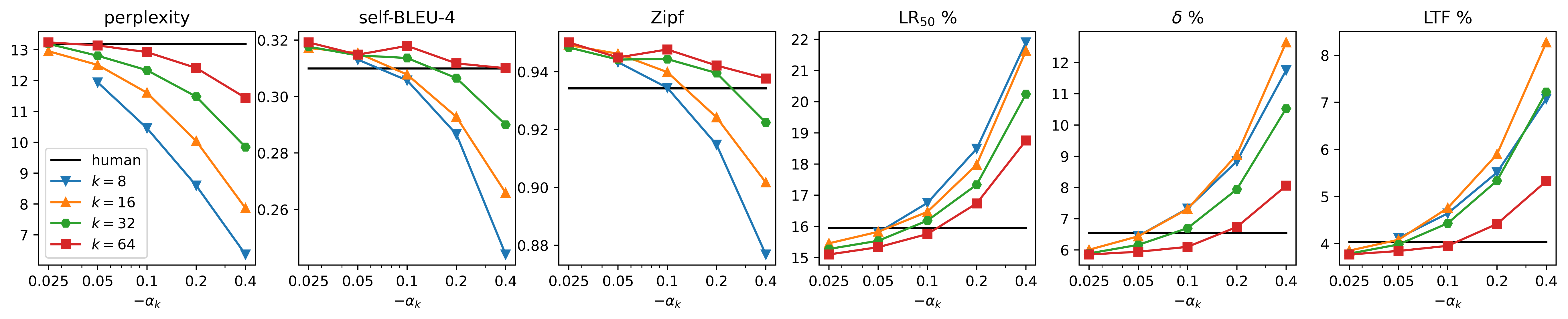}
\caption{Effect of $k$ and $\alpha_k$ on metrics from Table~\ref{tab:owt-gen}. The horizontal line marks the score of the human reference.}
\label{fig:owt-metrics}
\end{figure*}

The experiment in this section extends that of \citet{holtzman2020curious}. A selection of 5000 articles from WebText \citep{radford2019language} is taken as a reference corpus of human-written text. A language model (for us, GPT-2 Large) is prompted to generate text conditioned only on the \textit{first sentence} of each of these articles, up to a maximum of 200 tokens, yielding 5000 machine-generated texts. 

The human-written and machine-generated texts are compared by four automatic metrics: \textbf{perplexity} under the base LM, \textbf{self-BLEU-4} (\citet{zhu2018texygen}; the mean BLEU-4 score of a generated text with respect to all other generated texts as references), \textbf{Zipf coefficient} (the linear regression coefficient between log-rank and log-frequency of generated tokens) and \textbf{repetition} (the fraction of generated texts that end in a repeating sequence of tokens). It is desirable for a model and inference procedure to produce text that is as close as possible in these metrics to the human-written reference.

To measure long-range semantic coherence in the generated text, we define three new metrics:

\noindent\textbf{Long-range repetition (LR$_n$):} For a whole number $n$ and document $D$, let $S(D)$ be the number of distinct tokens in $D$, and let $R_n(D)$ be the number of distinct tokens for which the distance between their first and last occurrence in $D$ is at least $n$ positions. The long-range repetition score ${\rm LR}_n$ of a corpus $\{D_1,\dots,D_{5000}\}$ is a macro-average: \[{\rm LR}_n:=\frac{\sum_{i=1}^{5000}R_n(D_i)}{\sum_{i=1}^{5000}S(D_i)}.\]
This simple measure of lexical coherence favors repetition of words long after they are first used, but gives lower weight to documents that degenerate into repetition of a short span.

\noindent\textbf{Long-dependent token frequency (LTF):} A \textit{long-dependent token} is one to which the base LM assigns a likelihood of at least 20\% given its full context, but a likelihood of less than 5\% given only the 20 tokens of context preceding it. We compute the frequency of long-dependent tokens among all generated tokens.

\noindent\textbf{Long-short likelihood difference ($\delta$):} The mean difference in likelihoods assigned to tokens by the base LM conditioned on full context and conditioned on 20 tokens of context.


Although some choices of constants are needed to define \textbf{LTF} and $\delta$, we intend them to be intuitive summaries of long-range coherence in the absence of established metrics. In particular, 20 tokens is close to the length of one sentence in typical English text.

We sample 5000 document completions from GPT-2 Large following sampling procedures with a range of boosting schemes. We consider models of the form $f_k^{\alpha_k}f_{\max}^{1-\alpha_k}$, for $k\in\{8,16,32,64\}$ and $\alpha_k\in\{-0.4,-0.2,-0.1,-0.05,-0.025,0\}$. (Such a parametrization of boosting parameters was chosen to ensure that when the context has length less than $k$ -- or the distant context has very little effect on the next word -- the boosted model becomes equivalent to the untempered $f_{\rm max}$.) Top-$p$ truncation with $p=0.95$ is applied to all models.

\paragraph{Results.} Metrics of two of the best models, with $k=32,\alpha_k=-0.05$ and $k=64,\alpha_k=-0.1$, are shown in Table~\ref{tab:owt-gen}. In particular, the latter model generates text that is closer to the human reference, or equally close, to the pure top-$p$ sampling ($\alpha_k=0$) baseline in all metrics, with the greatest improvement seen in the coherence measures.

Fig.~\ref{fig:owt-metrics} shows the dependence of selected metrics on $k$ and $\alpha_k$. Coherence boosting brings all metrics closer to those of human text. As $k$ increases, the optimal $\alpha_k$ grows in magnitude. This is expected: the predictive effect of tokens more than $k$ positions away decreases with $k$ ($f_k$ approaches $f_{\rm max}$).

We also note that a simple sampling with temperature 0.9 performs better than top-$p$ sampling in most of the coherence metrics. This suggests that the improvements accomplished by top-$p$ truncation come at the cost of introducing a bias towards tokens that are predictable from a short context. Coherence boosting corrects this bias without sacrificing the gains in other measures. 

An example of human, top-$p$, and coherence boosting outputs is shown in Table~\ref{tab:owt-examples}.

\subsection{Dialog systems}
\label{sec:experiments_dialog}

\begin{table*}[t]
\centering
\scalebox{0.9}{
\begin{tabular}{ lccccccccc}
\toprule
& \multicolumn{2}{c}{NIST} & \multicolumn{2}{c}{BLEU} & & \multicolumn{3}{c}{diversity metrics} \\\cmidrule(lr){2-3}\cmidrule(lr){4-5}\cmidrule(lr){7-9}
Inference method & N-2 & N-4 & B-2 & B-4 & METEOR & Ent-4 & Dist-1 & Dist-2 & avg len \\\midrule
Beam ($b=10$) & 0.02 & 0.02 & {12.81} & {3.23} & 5.35 & 6.06 & 14.03 & 34.59 & 5.81 \\
Greedy          & 1.62 & 1.63 & 9.92 & 1.72 & 6.78 & 6.45 & 6.19 & 17.56 & 13.30 \\ \midrule
+ boost ($\alpha=-0.3$) & 0.72 & 0.73 & \textbf{13.82} & \textbf{3.53} & \textbf{6.91} & 8.54 & \textbf{16.81} & 49.35 & 9.75 \\
+ boost ($\alpha=-0.7$) & \textbf{1.78} & \textbf{1.79} & 6.33 & 0.94 & 5.55 & \textbf{9.78} & 28.00 & \textbf{72.46} & \textbf{16.63} \\
\midrule
Human           & 2.63 & 2.65 & 12.36 & 3.13 & 8.31 & 10.44 & 16.65 & 67.01 & 18.73 \\
\bottomrule
\end{tabular}
}
\vspace*{-1em}
\caption{Metrics of DialoGPT responses on DSTC7. Nearest-to-human values in each column are \textbf{bolded}.}
\label{tab:dialog-gen}
\end{table*}

This experiment is based on the Dialog System Technology Challenge 7 (DSTC7) \citep{galley2019grounded}, which benchmarks generation of dialog responses conditioned on one or more turns of conversation context. As a base model, we use DialoGPT \cite{zhang-etal-2020-dialogpt}, a GPT-2 Small variant that demonstrated strong results on this task. 

Dialog systems' responses to the 2208 conversation prompts\footnote{The DSTC7 evaluation data, scraped from Reddit, is undisclosed; we reacquire it using officially released code.} are scored against human-written reference responses (five for each example). Following \citet{zhang-etal-2020-dialogpt}, we use the $n$-gram overlap metrics \textbf{NIST} \citep{doddington2002automatic}, \textbf{BLEU} \citep{papineni-etal-2002-bleu}, and \textbf{METEOR} \citep{lavie-agarwal-2007-meteor}, as well as two intrinsic measures of $n$-gram diversity from \citet{li-etal-2016-diversity,zhang2018generating}:  \textbf{Distinct-$n$} and \textbf{Entropy-$n$}. It is desirable for a dialog system to reach scores close to those of the human responses in all metrics.

In addition to the decoding algorithms considered by \cite{zhang-etal-2020-dialogpt} -- beam search and greedy decoding -- we consider greedy decoding with a coherence boosting model. As long and short predictors, we use DialoGPT conditioned on the full conversation context and on \emph{only the (context-free) response generated so far}. That is, if the conversation context is $S$ and the text generated so far is $w_1\dots w_k$, then $w_{k+1}$ is predicted using the model $f_{\rm max}f_{k+1}^{\alpha}$, evaluated on the string $S\ \gq{\rm sep}\ w_1\dots w_k$, where $\gq{\rm sep}$ is the turn separator token. We consider $\alpha\in\{0,-0.1,\dots,-0.8\}$.

\paragraph{Results.} Table~\ref{tab:dialog-gen} shows the metrics of the boosting models that reach the peak average NIST and BLEU scores ($\alpha=-0.3$ and $\alpha=-0.7$). 
Increasing the magnitude of $\alpha$ leads to responses that are more relevant to the prompt (higher BLEU and NIST) and more diverse than those from greedy decoding. As $-\alpha$ grows large, the boosting model favors creative responses that are relevant to the prompt (high NIST), but simple responses that are common in the reference data become unlikely (low BLEU).\footnote{\citet{galley2019grounded} argue that NIST and diversity metrics are more informative measures than BLEU for multi-reference scoring, since BLEU favors systems that often produce responses with little relation to the prompt (e.g., ``I don't know'').} 

We observed that the responses with $\alpha=-0.7$, despite the superior metrics, are more likely to be ungrammatical and innovate words in an effort to use tokens relevant to the prompt. In practice, improving dialog systems with coherence boosting may require techniques to prevent these side effects, such as repetition penalties or relaxation of greedy decoding to low-temperature sampling.

Finally, we note that the learning of DialoGPT was initialized with a pretrained GPT-2 and uses GPT-2's end-of-text token as the turn separator. This choice may reduce DialoGPT's attention to past turns, as tokens \emph{preceding} the end-of-text token are never informative in GPT-2's training data.

\section{Experiments: Language understanding}
\label{sec:experiments_nlu}

\begin{table*}[t!]
\centering
\resizebox{1\linewidth}{!}{
\begin{tabular}{lcccrcccrcccr}
\toprule
            & \multicolumn{4}{c}{GPT-2 Small (125M)}                                & \multicolumn{4}{c}{GPT-2 XL (1.6B)}                                   & \multicolumn{4}{c}{GPT-3 175B}                                       \\ \cmidrule(lr){6-9}\cmidrule(lr){10-13}\cmidrule(lr){2-5}
            & $f_{\max}$ & $\alpha=-1$     & $\alpha=\alpha^*$             & \multicolumn{1}{c}{$\alpha^*$}      & $f_{\max}$ & $\alpha=-1$     & $\alpha=\alpha^*$             & \multicolumn{1}{c}{$\alpha^*$}     & $f_{\max}$     & $\alpha=-1$     & $\alpha=\alpha^*$             & \multicolumn{1}{c}{$\alpha^*$}      \\ \midrule
StoryCloze & 59.91      & \textbf{64.78} & 64.24          & \textit{$-$1.02} & 67.56      & 75.09          & \textbf{76.75} & \textit{$-$0.69} & 79.16          & 82.90          & \textbf{86.85} & \textit{$-$0.64} \\
HellaSwag   & 28.92      & 30.99          & \textbf{31.84} & \textit{$-$0.90} & 40.00      & 42.60          & \textbf{47.66} & \textit{$-$0.78} & 59.18          & 62.66          & \textbf{72.35} & \textit{$-$0.76} \\
COPA        
& {62.00}      & 56.00          & \textbf{64.00} & \textit{$-$0.69}  
& 73.00      & 70.00          & \textbf{77.00} & \textit{$-$0.44} 
& 93.00          & 87.00          & \textbf{94.00} & \textit{$-$0.52} \\ \midrule
CsQA        & 29.48      & 42.26          & \textbf{43.16} & \textit{$-$0.81} & 37.84      & 50.45          & \textbf{52.91} & \textit{$-$0.75} & 61.10          & 67.98          & \textbf{70.43} & \textit{$-$0.68} \\
OBQA        & 11.20      & 30.60          & \textbf{40.80} & \textit{$-$1.62} & 15.60      & 38.40          & \textbf{47.00} & \textit{$-$1.88} & 28.00          & 52.20          & \textbf{52.60} & \textit{$-$1.09} \\
ARC-E       & 43.81      & 42.09          & \textbf{46.00} & \textit{$-$0.34} & 58.29      & 51.43          & \textbf{60.31} & \textit{$-$0.36} & 76.22          & 69.19          & \textbf{78.32} & \textit{$-$0.44} \\
ARC-C       & 19.03      & 26.11          & \textbf{29.10} & \textit{$-$4.19} & 25.00      & 33.53          & \textbf{34.39} & \textit{$-$1.14} & 43.94          & \textbf{50.60} & 49.23          & \textit{$-$1.08} \\
PIQA        & 62.89      & 57.45          & \textbf{63.44} & \textit{$-$0.61} & 70.84      & 60.45          & \textbf{71.49} & \textit{$-$0.43} & 79.27          & 66.32          & \textbf{78.94} & \textit{$-$0.60} \\ \midrule
SST2        & 65.68      & 74.74          & \textbf{82.32} & \textit{$-$2.22} & 86.38      & 84.51          & \textbf{86.93} & \textit{$-$0.09} & 86.16          & 88.14          & \textbf{89.84} & \textit{$-$0.54} \\
SST5        & 25.93      & \textbf{30.90} & \textbf{30.90} & \textit{$-$1.20} & 28.69      & \textbf{38.73} & 36.92          & \textit{$-$1.69} & 31.22          & 34.75          & \textbf{38.51} & \textit{$-$1.39} \\
AGNews      & 58.55      & 60.78          & \textbf{62.20} & \textit{$-$0.62} & 67.17      & 67.43          & \textbf{68.26} & \textit{$-$0.40} & 71.66          & 71.74          & \textbf{71.75} & \textit{0.16}  \\
TREC        & 23.40      & 29.60          & \textbf{32.20} & \textit{$-$0.80} & 23.40      & 27.40          & \textbf{40.00} & \textit{$-$0.79} & 52.40          & 47.00          & \textbf{56.00} & \textit{$-$0.56} \\
BoolQ       & 49.36      & 58.07          & \textbf{62.14} & \textit{$-$3.04} & 62.14      & \textbf{63.46} & 63.21          & \textit{$-$0.64} & 71.56          & \textbf{73.70} & 72.69          & \textit{$-$0.39} \\ \midrule
RTE         
& 51.26      & 49.82 & \textbf{53.79}          & \textit{$-$0.30}  
& \textbf{49.10}      & 48.74 & \textbf{49.10}          & \textit{0.90}  
& 55.96 & 57.40 & \textbf{60.29}          & \textit{$-$0.60}  \\
CB         
& 12.50      & {23.21} & \textbf{48.21} & \textit{$-$2.40} 
& 30.36      & 51.79          & \textbf{66.07} & \textit{$-$1.90}  
& {5.36} & 25.00          & \textbf{28.57} & \textit{$-$1.91}  \\  \midrule
Average & 40.26 & 45.16 & \textbf{50.29} & \textit{$-$1.39} & 49.02 & 53.60 & \textbf{58.53} & \textit{$-$0.74} & 59.61 & 62.44 & \textbf{66.69} & \textit{$-$0.73}\\
\bottomrule
\end{tabular}
}

\vspace*{-1em}\caption{Testing accuracy (\%) of three representative GPT models on multiple-choice tasks. The first column for each model is the full-context model, the second is our model only when $\alpha=-1$ (a baseline), and the third column is our model with the optimal $\alpha$ chosen on a validation set. The fourth column shows this optimal value of $\alpha$.}
\label{tab:multichoice-main}
\end{table*}

We evaluate coherence boosting on zero-shot language understanding and inference tasks, where examples are expected to be highly coherent.

We study 15 datasets in 5 categories of tasks.
\textbf{(1)} \textbf{Cloze tasks}: 
\textit{StoryCloze}~\citep{mostafazadeh-etal-2016-corpus}, \textit{HellaSwag}~\citep{zellers-etal-2019-hellaswag}, and \textit{COPA}~\citep{roemmele2011choice}.
\textbf{(2)} \textbf{Question answering}: 
\textit{CommonsenseQA} (CsQA)~\citep{talmor-etal-2019-commonsenseqa}, \textit{OpenBookQA} (OBQA)~\citep{mihaylov-etal-2018-suit}, \textit{ARC Easy / Challenge} (ARC-E/C) \citep{clark2018think}, and \textit{PIQA}~\citep{bisk2020piqa}.
\textbf{(3)} \textbf{Text classification}:
\textit{SST-2/5}~\citep{socher-etal-2013-recursive}, \textit{TREC}~\citep{voorhees2000building}, \textit{AGNews}~\citep{zhang2015character}.
\textbf{(4)} \textbf{Natural language inference}:
\textit{RTE}~\citep{dagan2005pascal}, \textit{CB}~\citep{de2019commitmentbank}, and \textit{BoolQ}~\cite{clark-etal-2019-boolq}.
(5) \textbf{Fact knowledge retrieval}:
\textit{LAMA}~\citep{petroni-etal-2019-language}.

All tasks except LAMA are formulated as multiple-choice problems. We convert text classification and inference tasks to multiple-choice tasks by choosing meaningful answer words, e.g., ``True''/``False''. The prediction is made by selecting the choice with the highest LM likelihood. 

For in-context learning of GPT models, prompt formats greatly impact performance. We follow previous work~\citep{brown2020language, zhao2021calibrate, holtzman2021surface} to create natural prompts to enlarge the effectiveness of in-context learning, but we do not aim to optimize the full and context-free prompt format: our goal is to evaluate coherence boosting models with a fixed prompt.
The prompt formats we use are listed in Table~\ref{tab:prompt_format}. As described in \S\ref{sec:cb}, within each prompt we identify a \emph{premise-free context}, which is used as the context for the short-range model in coherence boosting.

\begin{figure}[h]
    \centering
    \includegraphics[width=.45\textwidth]{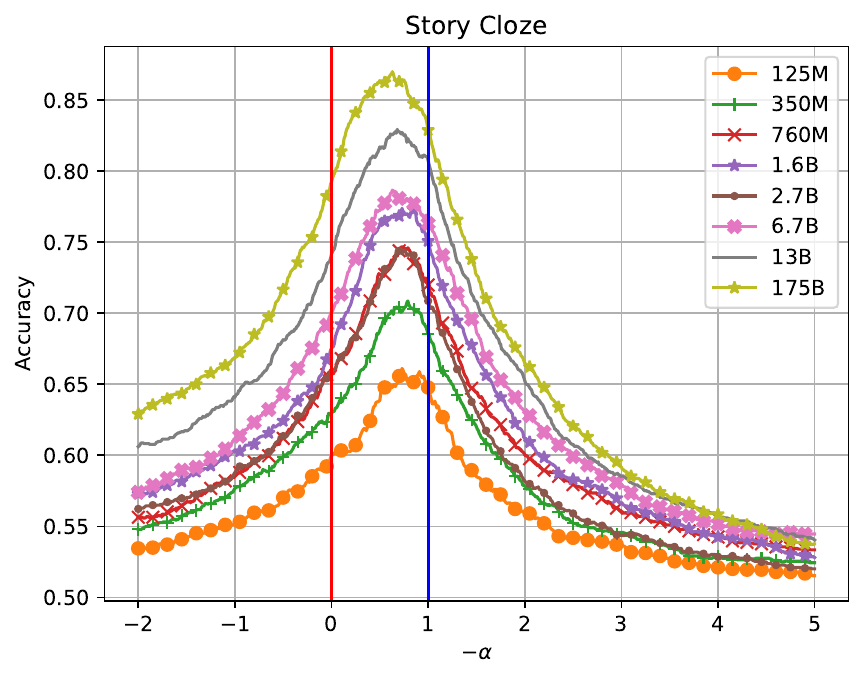}
    \vspace*{-1em}
    \caption{Model comparison for the StoryCloze task. The red line $\alpha=0$ indicates the base model, and the blue line $\alpha=-1$ is an unconditional normalization. See Figs.~\ref{fig:multichoice_1} and \ref{fig:multichoice_2} for plots for other tasks, and note that they do not all have the same shape.}
    \label{fig:multichoice_alpha_storcycloze}
\end{figure}

For each dataset, we pick the optimal value $\alpha^*$ of the parameter $\alpha$ on the validation set and report the accuracy on testing set. (If no testing set is publicly available, we choose $\alpha$ on a subset of the training set and report the final number on the validation set.) Across all experiments, we do not put any few-shot examples in the prompt.

For the knowledge retrieval task, we follow \citet{zhao2021calibrate}'s data split of LAMA and evaluate GPT models on facts whose missing answers are at the end of the sentence (to fit the nature of autoregressive language models). We limit the prompt length to be larger than 5 tokens and rerun the model from \citet{zhao2021calibrate} on the new data.

\paragraph{Results: Multiple-choice tasks.}

Results of three representative base models on all multiple-choice tasks are presented in Table~\ref{tab:multichoice-main}. (Results for all models are in Tables~\ref{tab:multichoice-gpt3} and \ref{tab:multichoice-gpt2}.) We compare our best model with two baselines, $\alpha=0$ ($f_{\max}$) and $\alpha=-1$. The former one is the original full-context model, while the latter is, for most tasks, a form of unconditional probability normalization as performed by \citet{brown2020language, holtzman2021surface}. We also compare our best model with other inference methods~\citep{holtzman2021surface, min2021noisy} in Tables~\ref{tab:multichoice-pmi-gpt3} and \ref{tab:multichoice-pmi-gpt2}.

By comparing the third column with the first two columns within each model in Table~\ref{tab:multichoice-main}, we can see that our method with the selected $\alpha$ generally improves the accuracy on all tasks. Some of the improvements are dramatic, where boosted GPT-2 Small outperforms GPT-2 XL's base model (e.g., CsQA, OBQA, ARC-C) and is even comparable with GPT-3 175B's base model (e.g., SST-2, SST-5, RTE). We make similar conclusions when comparing coherence boosting with other inference methods in Tables~\ref{tab:multichoice-pmi-gpt3} and \ref{tab:multichoice-pmi-gpt2}.

We observe that the optimal $\alpha$ depends on tasks and models (fourth column within each model), which means that $\alpha$ cannot be heuristically set to $0$ or $-1$ as in past work. This finding suggests the necessity of searching for an optimal $\alpha$. We visualize the accuracy curve by varying $\alpha$ in the testing set of all datasets. We show the curve for StoryCloze in Fig.~\ref{fig:multichoice_alpha_storcycloze} and present similar figures for all tasks in Figs.~\ref{fig:multichoice_1} and \ref{fig:multichoice_2}.

\begin{table*}[t!]
\centering
\resizebox{0.75\linewidth}{!}{

\begin{tabular}{lcccccccc}
\toprule
& \multicolumn{4}{c}{GPT-2} & \multicolumn{4}{c}{GPT-3} \\\cmidrule(lr){2-5}\cmidrule(lr){6-9}
           & 125M & 350M & 760M & 1.6B & 2.7B & 6.7B & 13B  & 175B \\ \midrule
$f_{\max}$  & 8.48  & 14.78 & 13.88 & 14.29 & 17.33 & 19.42 & 22.06 & 26.76 \\
\citet{zhao2021calibrate}        & 17.45 & 22.87 & 23.90 & 23.97 & 26.30 & 30.57 & 31.96 & 34.78  \\
CB  ($\alpha_k=\alpha_k^*$)       & 19.85 & 22.87 & 25.74 & 25.43 & 28.75 & 32.25 & 35.02 & 37.57 \\ 
\midrule
$\alpha_k^*$   & $-$0.5  & $-$0.5  & $-$0.5  & $-$0.5  & $-$0.5  & $-$0.5  & $-$0.5  & $-$0.4  \\
$k^*$       & 1    & 2    & 3    & 3    & 1    & 1    & 1    & 2   \\ \bottomrule
\end{tabular}

}
\caption{Accuracies (\%) of GPT models on LAMA.}
\label{tab:lama}
\end{table*}

Consistent with the results on LAMBADA (\S\ref{sec:lambada}), the optimal $\alpha$ is usually negative, and its absolute value tends to decrease with the model size. We selected the optimal $\alpha$ by the validation set, but future work may explore automatic and adaptive methods for setting this parameter. Notice that all experiments required only a \emph{single pass} through the data to compute answer likelihoods conditioned on full and premise-free contexts -- no iterative gradient-based finetuning was applied.

\paragraph{Results: Knowledge retrieval.}

Unlike LAMBADA, where long contexts are required for inferring the last word, LAMA contains much shorter sentences for knowledge facts, i.e., (subject, relation, object). A recent study \citep{cao2021knowledgeable} shows that the prediction is biased by the relation in the short context, i.e., the answer to a prompt (e.g., ``Dante was born in \_\_\_'') can be induced by the relation (``was born in'') without the subject. Coherence boosting mitigates the influence of those short contexts by making the prediction dependent on a longer context containing the subject.

We present results for all models on LAMA in Table~\ref{tab:lama}. We also compare our model with contextual calibration (CC)  \citep{zhao2021calibrate}, which processes the LM's output probabilities with a log-linear model.
\footnote{Note that CC applies a log-linear model to the \emph{probability} domain, not the logit domain, which does not have an information-theoretic interpretation.} 
Coherence boosting with the selected $\alpha$ and $k$ outperforms both the base model and CC by significant margins.

\section{Extensions and future work}


We suggest three promising research directions:

\paragraph{Coherence tuning.} The need to evaluate the base LM with multiple contexts in coherence boosting introduces cost and complexity at inference time. It may be desirable instead to modify the weights of the base model to improve long-range coherence properties. In \S\ref{sec:coherence_tuning}, we describe a `self-tuning' algorithm that achieves this \emph{without training on any data created for this purpose}.

\paragraph{New domains and architectures.} 

In this paper, we mainly considered coherence boosting with decoder-only Transformer LMs trained on generic text, but future work should consider other architectures and target domains. In \S\ref{sec:summarization}, we give preliminary results on the \emph{text summarization} domain.

Although we expect recency bias to be less pronounced in LMs that use separate attention modules to process the prompt and the output -- such as encoder-decoder models for translation or summarization -- procedures inspired by coherence boosting may prove effective in domains where a strong causal link between prompt and output is known to exist. Such domains include language generation conditioned on structured data~\citep{yao-etal-2020-heterogeneous, mager-etal-2020-gpt, moosavi2021learning} and model-guided reasoning in formal languages, such as proof or program synthesis~\cite{gptf, chen2021evaluating, alphacode}.

\paragraph{Efficient search proposals.} Procedures that force LMs to be more focused on a prompt, or a specific part of it, when generating or ranking tokens can benefit algorithms that search for combinations of words through sampling. It would be interesting to use coherence boosting in non-autoregressive text generation algorithms, such as to accelerate the mixing of MCMC methods for constrained text generation \citep{miao2019cgmh,zhang-etal-2020-language-generation,whitammer-etal-2021-studying}.

\section{Conclusion}

We have illustrated the hyposensitivity of pretrained language models to long-range context and proposed a simple inference-time remedy. We hope to see coherence boosting used as a simple alternative or complement to finetuning procedures in zero-shot applications of pretrained LMs.

\section*{Acknowledgments}

The authors are grateful to Sudha Rao, Matt Richardson, and Huan Sun for valuable discussions about this project. We thank the anonymous reviewers for their comments and suggestions.

\section*{Ethics statement}

We hope and expect to see a nonnegative net societal impact from better text generation and ranking algorithms in general and from this work in particular. As we have shown, there is room to improve the inference procedures used with small language models, which incur lower costs than training and evaluation of large models. However, researchers should bear in mind the risks and potential misuse of automatic generation of long-form text.

\bibliography{anthology,references}

@article{zhang2018generating,
  title={Generating Informative and Diverse Conversational Responses via Adversarial Information Maximization},
  author={Zhang, Yizhe and Galley, Michel and Gao, Jianfeng and Gan, Zhe and Li, Xiujun and Brockett, Chris and Dolan, William B.},
  journal={Neural Information Processing Systems (NeurIPS)},
  year={2018}
}

@article{holtzman2020curious,
  title={The Curious Case of Neural Text Degeneration},
  author={Holtzman, Ari and Buys, Jan and Du, Li and Forbes, Maxwell and Choi, Yejin},
  journal={International Conference on Learning Representations (ICLR)},
  year={2019}
}

@article{zhao2021calibrate,
  title = 	 {Calibrate Before Use: Improving Few-shot Performance of Language Models},
  author =       {Zhao, Zihao and Wallace, Eric and Feng, Shi and Klein, Dan and Singh, Sameer},
  journal = 	 {International Conference on Machine Learning (ICML)},
  year = 	 {2021},
}

@article{brown2020language,
 author = {Brown, Tom and Mann, Benjamin and Ryder, Nick and Subbiah, Melanie and Kaplan, Jared D and Dhariwal, Prafulla and Neelakantan, Arvind and Shyam, Pranav and Sastry, Girish and Askell, Amanda and Agarwal, Sandhini and Herbert-Voss, Ariel and Krueger, Gretchen and Henighan, Tom and Child, Rewon and Ramesh, Aditya and Ziegler, Daniel and Wu, Jeffrey and Winter, Clemens and Hesse, Chris and Chen, Mark and Sigler, Eric and Litwin, Mateusz and Gray, Scott and Chess, Benjamin and Clark, Jack and Berner, Christopher and McCandlish, Sam and Radford, Alec and Sutskever, Ilya and Amodei, Dario},
 journal = {Neural Information Processing Systems (NeurIPS)},
 title = {Language Models are Few-Shot Learners},
 year = {2020}
}

@article{roemmele2011choice,
  title={Choice of plausible alternatives: An evaluation of commonsense causal reasoning},
  author={Roemmele, Melissa and Bejan, Cosmin Adrian and Gordon, Andrew S.},
  journal={AAAI Spring Symposium Series},
  year={2011}
}

@article{bengio2003neural,
author = {Bengio, Yoshua and Ducharme, R\'{e}jean and Vincent, Pascal and Janvin, Christian},
title = {A Neural Probabilistic Language Model},
year = {2003},
issue_date = {3/1/2003},
volume = {3},
journal = {Journal of Machine Learning Research},
month = mar,
pages = {1137–1155},
numpages = {19}
}

@article{galley2019grounded,
  title={Grounded response generation task at {DSTC7}},
  journal={Dialog System Technology Challenges 7 (AAAI workshop)},
  author={Galley, Michel and Brockett, Chris and  Gao, Xiang and Gao, Jianfeng and Dolan, William B.},
  year={2019},
}

@article{radford2019language,
  title={Language models are unsupervised multitask learners},
  author={Radford, Alec and Wu, Jeffrey and Child, Rewon and Luan, David and Amodei, Dario and Sutskever, Ilya},
  year={2019}
}

@article{press2021train,
      title={Train Short, Test Long: Attention with Linear Biases Enables Input Length Extrapolation}, 
      author={Ofir Press and Noah A. Smith and Mike Lewis},
      year={2022},
      journal={International Conference on Learning Representations (ICLR)},
}

@inproceedings{meister-cotterell-2021-language,
    title = "Language Model Evaluation Beyond Perplexity",
    author = "Meister, Clara  and
      Cotterell, Ryan",
    booktitle = "Proceedings of the 59th Annual Meeting of the Association for Computational Linguistics and the 11th International Joint Conference on Natural Language Processing (Volume 1: Long Papers)",
    month = aug,
    year = "2021",
    address = "Online",
    publisher = "Association for Computational Linguistics",
    url = "https://aclanthology.org/2021.acl-long.414",
    doi = "10.18653/v1/2021.acl-long.414",
    pages = "5328--5339"
}

@article{beltagy2020longformer,
  title={Longformer: The Long-Document Transformer},
  author={Iz Beltagy and Matthew E. Peters and Arman Cohan},
  journal={arXiv preprint arXiv:2004.05150},
  year={2020},
}

@article{hochreiter1997long,
    author = {Hochreiter, Sepp and Schmidhuber, J\"{u}rgen},
    title = {Long Short-Term Memory},
    year = {1997},
    publisher = {MIT Press},
    address = {Cambridge, MA, USA},
    volume = {9},
    number = {8},
    journal = {Neural Computation},
    month = {11},
    pages = {1735--1780},
    numpages = {46}
}

@article{bahdanau2015neural,
    author = {Bahdanau, Dzmitry and Cho, Kyunghyun and Bengio, Yoshua},
    year = {2015},
    journal={International Conference on Learning Representations (ICLR)},
    title = {Neural Machine Translation by Jointly Learning to Align and Translate}
}

@article{vaswani2017attention,
    author = {Vaswani, Ashish and Shazeer, Noam and Parmar, Niki and Uszkoreit, Jakob and Jones, Llion and Gomez, Aidan N. and Kaiser, Lukasz and Polosukhin, Illia},
    year = {2017},
    journal={Neural Information Processing Systems (NIPS)},
    title = {Attention is All You Need}
}

@article{zaheer2020big,
  title={Big Bird: Transformers for Longer Sequences},
  author={Zaheer, Manzil and Guruganesh, Guru and Dubey, Kumar Avinava and Ainslie, Joshua and Alberti, Chris and Ontanon, Santiago and Pham, Philip and Ravula, Anirudh and Wang, Qifan and Yang, Li and others},
  journal={Neural Information Processing Systems (NeurIPS)},
  year={2020}
}

@article{roy-etal-2021-efficient,
    title = "Efficient Content-Based Sparse Attention with Routing Transformers",
    author = "Roy, Aurko  and
      Saffar, Mohammad  and
      Vaswani, Ashish  and
      Grangier, David",
    journal = "Transactions of the Association for Computational Linguistics",
    volume = "9",
    year = "2021",
    url = "https://aclanthology.org/2021.tacl-1.4",
    doi = "10.1162/tacl_a_00353",
    pages = "53--68"
}

@inproceedings{doddington2002automatic,
  title={Automatic evaluation of machine translation quality using n-gram co-occurrence statistics},
  author={George R. Doddington},
  booktitle={Human Language Technology Research},
  year={2002}
}

@inproceedings{voorhees2000building,
  title={Building a question answering test collection},
  author={Voorhees, Ellen M and Tice, Dawn M},
  booktitle={Proceedings of the 23rd annual international ACM SIGIR conference on Research and Development in Information Retrieval},
  pages={200--207},
  year={2000}
}

@inproceedings{de2019commitmentbank,
  title={The CommitmentBank: Investigating projection in naturally occurring discourse},
  author={De Marneffe, Marie-Catherine and Simons, Mandy and Tonhauser, Judith},
  booktitle={proceedings of Sinn und Bedeutung},
  volume={23},
  number={2},
  pages={107--124},
  year={2019}
}

@inproceedings{dagan2005pascal,
  title={The pascal recognising textual entailment challenge},
  author={Dagan, Ido and Glickman, Oren and Magnini, Bernardo},
  booktitle={Machine Learning Challenges Workshop},
  pages={177--190},
  year={2005},
  organization={Springer}
}

@article{zhang2015character,
  title={Character-level convolutional networks for text classification},
  author={Zhang, Xiang and Zhao, Junbo and LeCun, Yann},
  journal={Neural Information Processing Systems (NIPS)},
  year={2015}
}

@article{clark2018think,
  title={{Think you have solved question answering? Try ARC, the AI2 reasoning challenge}},
  author={Clark, Peter and Cowhey, Isaac and Etzioni, Oren and Khot, Tushar and Sabharwal, Ashish and Schoenick, Carissa and Tafjord, Oyvind},
  journal={arXiv preprint arXiv:1803.05457},
  year={2018}
}

@article{bisk2020piqa,
  title={Piqa: Reasoning about physical commonsense in natural language},
  author={Bisk, Yonatan and Zellers, Rowan and Gao, Jianfeng and Choi, Yejin and others},
  journal={Association for the Advancement of Artificial Intelligence (AAAI)},
  year={2020}
}

@inproceedings{cao2021knowledgeable,
    title = "Knowledgeable or Educated Guess? Revisiting Language Models as Knowledge Bases",
    author = "Cao, Boxi  and
      Lin, Hongyu  and
      Han, Xianpei  and
      Sun, Le  and
      Yan, Lingyong  and
      Liao, Meng  and
      Xue, Tong  and
      Xu, Jin",
    booktitle = "Proceedings of the 59th Annual Meeting of the Association for Computational Linguistics and the 11th International Joint Conference on Natural Language Processing (Volume 1: Long Papers)",
    month = aug,
    year = "2021",
    address = "Online",
    publisher = "Association for Computational Linguistics",
    url = "https://aclanthology.org/2021.acl-long.146",
    doi = "10.18653/v1/2021.acl-long.146",
    pages = "1860--1874",
}

@article{min2021noisy,
  title={Noisy channel language model prompting for few-shot text classification},
  author={Min, Sewon and Lewis, Mike and Hajishirzi, Hannaneh and Zettlemoyer, Luke},
  journal={arXiv preprint arXiv:2108.04106},
  year={2021}
}

@article{friedman2000additive,
author = {Friedman, Jerome and Hastie, Trevor and Tibshirani, Robert},
year = {2000},
month = {04},
pages = {337-407},
title = {Additive Logistic Regression: A Statistical View of Boosting},
volume = {28},
journal = {The Annals of Statistics},
}

@article{zhu2018texygen,
author = {Zhu, Yaoming and Lu, Sidi and Zheng, Lei and Guo, Jiaxian and Zhang, Weinan and Wang, Jun and Yu, Yong},
title = {Texygen: A Benchmarking Platform for Text Generation Models},
year = {2018},
journal = {ACM SIGIR Conference on Research and Development in Information Retrieval}
}

@article{miao2019cgmh,
 author={Miao, Ning and  Zhou, Hao and Mou, Lili and Yan, Rui and and Li Lei},
 title={{CGMH}: Constrained sentence generation by {Metropolis-Hastings} sampling},
 year={2019},
 journal={Association for the Advancement of Artificial Intelligence (AAAI)}
}

@inproceedings{sun-etal-2021-long,
    title = "Do Long-Range Language Models Actually Use Long-Range Context?",
    author = "Sun, Simeng  and
      Krishna, Kalpesh  and
      Mattarella-Micke, Andrew  and
      Iyyer, Mohit",
    booktitle = "Proceedings of the 2021 Conference on Empirical Methods in Natural Language Processing",
    month = nov,
    year = "2021",
    address = "Online and Punta Cana, Dominican Republic",
    publisher = "Association for Computational Linguistics",
    url = "https://aclanthology.org/2021.emnlp-main.62",
    pages = "807--822",
}

@inproceedings{holtzman2021surface,
    title = "Surface Form Competition: Why the Highest Probability Answer Isn{'}t Always Right",
    author = "Holtzman, Ari  and
      West, Peter  and
      Shwartz, Vered  and
      Choi, Yejin  and
      Zettlemoyer, Luke",
    booktitle = "Proceedings of the 2021 Conference on Empirical Methods in Natural Language Processing",
    month = nov,
    year = "2021",
    address = "Online and Punta Cana, Dominican Republic",
    publisher = "Association for Computational Linguistics",
    url = "https://aclanthology.org/2021.emnlp-main.564",
    pages = "7038--7051",
}

@inproceedings{schick-schutze-2021-exploiting,
    title = "Exploiting Cloze-Questions for Few-Shot Text Classification and Natural Language Inference",
    author = {Schick, Timo  and
      Sch{\"u}tze, Hinrich},
    booktitle = "Proceedings of the 16th Conference of the European Chapter of the Association for Computational Linguistics: Main Volume",
    month = apr,
    year = "2021",
    address = "Online",
    publisher = "Association for Computational Linguistics",
    url = "https://aclanthology.org/2021.eacl-main.20",
    doi = "10.18653/v1/2021.eacl-main.20",
    pages = "255--269",
}

@article{sanh-etal-2022-multitask,
title={Multitask Prompted Training Enables Zero-Shot Task Generalization},
author={Victor Sanh and Albert Webson and Colin Raffel and Stephen Bach and Lintang Sutawika and Zaid Alyafeai and Antoine Chaffin and Arnaud Stiegler and Arun Raja and Manan Dey and M Saiful Bari and Canwen Xu and Urmish Thakker and Shanya Sharma Sharma and Eliza Szczechla and Taewoon Kim and Gunjan Chhablani and Nihal Nayak and Debajyoti Datta and Jonathan Chang and Mike Tian-Jian Jiang and Han Wang and Matteo Manica and Sheng Shen and Zheng Xin Yong and Harshit Pandey and Rachel Bawden and Thomas Wang and Trishala Neeraj and Jos Rozen and Abheesht Sharma and Andrea Santilli and Thibault Fevry and Jason Alan Fries and Ryan Teehan and Teven Le Scao and Stella Biderman and Leo Gao and Thomas Wolf and Alexander M Rush},
journal={International Conference on Learning Representations (ICLR)},
year={2022}
}

@inproceedings{narayan2018don,
  title={Don’t Give Me the Details, Just the Summary! Topic-Aware Convolutional Neural Networks for Extreme Summarization},
  author={Narayan, Shashi and Cohen, Shay B and Lapata, Mirella},
  booktitle={Proceedings of the 2018 Conference on Empirical Methods in Natural Language Processing},
  pages={1797--1807},
  year={2018}
}

@inproceedings{zhong2020extractive,
  title={Extractive Summarization as Text Matching},
  author={Zhong, Ming and Liu, Pengfei and Chen, Yiran and Wang, Danqing and Qiu, Xipeng and Huang, Xuan-Jing},
  booktitle={Proceedings of the 58th Annual Meeting of the Association for Computational Linguistics},
  pages={6197--6208},
  year={2020}
}

@inproceedings{lewis2020bart,
  title={BART: Denoising Sequence-to-Sequence Pre-training for Natural Language Generation, Translation, and Comprehension},
  author={Lewis, Mike and Liu, Yinhan and Goyal, Naman and Ghazvininejad, Marjan and Mohamed, Abdelrahman and Levy, Omer and Stoyanov, Veselin and Zettlemoyer, Luke},
  booktitle={Proceedings of the 58th Annual Meeting of the Association for Computational Linguistics},
  pages={7871--7880},
  year={2020}
}

@inproceedings{zhang2020pegasus,
  title={Pegasus: Pre-training with extracted gap-sentences for abstractive summarization},
  author={Zhang, Jingqing and Zhao, Yao and Saleh, Mohammad and Liu, Peter},
  booktitle={International Conference on Machine Learning},
  pages={11328--11339},
  year={2020},
  organization={PMLR}
}

@inproceedings{howcroft-etal-2020-twenty,
    title = "Twenty Years of Confusion in Human Evaluation: {NLG} Needs Evaluation Sheets and Standardised Definitions",
    author = "Howcroft, David M.  and
      Belz, Anya  and
      Clinciu, Miruna-Adriana  and
      Gkatzia, Dimitra  and
      Hasan, Sadid A.  and
      Mahamood, Saad  and
      Mille, Simon  and
      van Miltenburg, Emiel  and
      Santhanam, Sashank  and
      Rieser, Verena",
    booktitle = "Proceedings of the 13th International Conference on Natural Language Generation",
    month = dec,
    year = "2020",
    address = "Dublin, Ireland",
    publisher = "Association for Computational Linguistics",
    url = "https://aclanthology.org/2020.inlg-1.23",
    pages = "169--182",
}

@inproceedings{belz-etal-2020-disentangling,
    title = "Disentangling the Properties of Human Evaluation Methods: A Classification System to Support Comparability, Meta-Evaluation and Reproducibility Testing",
    author = "Belz, Anya  and
      Mille, Simon  and
      Howcroft, David M.",
    booktitle = "Proceedings of the 13th International Conference on Natural Language Generation",
    month = dec,
    year = "2020",
    address = "Dublin, Ireland",
    publisher = "Association for Computational Linguistics",
    url = "https://aclanthology.org/2020.inlg-1.24",
    pages = "183--194",
}

@article{chen2021evaluating,
  title={Evaluating Large Language Models Trained on Code},
  author={Chen, Mark and Tworek, Jerry and Jun, Heewoo and Yuan, Qiming and de Oliveira Pinto, Henrique Ponde and Kaplan, Jared and Edwards, Harrison and Burda, Yuri and Joseph, Nicholas and Brockman, Greg and others},
  year={2021},
  journal={arXiv preprint 2107.03374}
}

@misc{alphacode,
    title={Competition-Level Code Generation with AlphaCode},
    author={Li, Yujia and Choi, David and Chung, Junyoung and Kushman, Nate and
    Schrittwieser, Julian and Leblond, Rémi and Eccles, Tom and
    Keeling, James and Gimeno, Felix and Dal Lago, Agustin and
    Hubert, Thomas and Choy, Peter and de Masson d'Autume, Cyprien and
    Babuschkin, Igor and Chen, Xinyun and Huang, Po-Sen and Welbl, Johannes and
    Gowal, Sven and Cherepanov, Alexey and Molloy, James and
    Mankowitz, Daniel and Sutherland Robson, Esme and Kohli, Pushmeet and
    de Freitas, Nando and Kavukcuoglu, Koray and Vinyals, Oriol},
    year={2022},
    month={Feb}}

@article{gptf,
title={Generative Language Modeling for Automated Theorem Proving},
author={Polu, Stanislas and Sutskever, Ilya},
journal={arXiv preprint 2009.03393},
year={2020}
}

@article{moosavi2021learning,
  title={Learning to reason for text generation from scientific tables},
  author={Moosavi, Nafise Sadat and R{\"u}ckl{\'e}, Andreas and Roth, Dan and Gurevych, Iryna},
  journal={arXiv preprint arXiv:2104.08296},
  year={2021}
}

@inproceedings{whitammer-etal-2021-studying,
    title = "Studying word order through iterative shuffling",
    author = "Whitammer, Esmeralda S.  and
      Lanka, Sameera  and
      Goel, Pranav  and
      Jojic, Nebojsa",
    editor = "Moens, Marie-Francine  and
      Huang, Xuanjing  and
      Specia, Lucia  and
      Yih, Scott Wen-tau",
    booktitle = "Proceedings of the 2021 Conference on Empirical Methods in Natural Language Processing",
    month = nov,
    year = "2021",
    address = "Online and Punta Cana, Dominican Republic",
    publisher = "Association for Computational Linguistics",
    url = "https://aclanthology.org/2021.emnlp-main.809/",
    doi = "10.18653/v1/2021.emnlp-main.809",
    pages = "10351--10366"
}
\bibliographystyle{style/acl_natbib}

\appendix

\onecolumn

\counterwithin{figure}{section}
\counterwithin{table}{section}

\section{On multi-objective training and log-linear weights}
\label{sec:derivation}

The section extends the discussion in \S\ref{sec:why_cb}.

Recall that the language model $f$ is trained on the multi-objective loss (\ref{eqn:criterion_scalarization}):
\[
\sum_{k=1}^M\lambda_k\underbrace{{\EE_{w_1\dots w_{M+1}\in\DDD}\left[-\log f_k(w_{M+1}|w_1,\dots,w_M;\theta)\right]}}_{\LLL_k(\theta)},\quad\lambda_k=\frac1M.
\]
As we saw in the main text, the scalarization weights $\lambda_k$ are uniform as a consequence of the training regime. However, evaluation procedures effectively give nonuniform weight to the $M$ prediction losses.

\paragraph{Some vector calculus.} Denote by $\hat\theta(\llambda)$ a local optimum of the above optimization problem for general linear combination weights $\llambda=(\lambda_1,\dots,\lambda_M)$. Under suitable regularity conditions, the gradient of the combined loss vanishes:
\begin{equation}
\left.\sum_k\lambda_k\frac{\partial\LLL_k(\theta)}{\partial\theta}\right|_{\theta=\hat\theta(\llambda)}=\boldsymbol{0}.
\label{eqn:derivative_alignment}
\end{equation}
Assuming the Hessian $\AA$ of the optimization criterion $\sum_k\lambda_k\LLL_k(\theta)$ is nonsingular, we can implicitly differentiate (\ref{eqn:derivative_alignment}) with respect to $\llambda$ to obtain the matrix derivative
\begin{equation}
    \frac{\partial\hat\theta(\llambda)}{\partial\llambda}
    =-\left.\AA^{-1}\frac{\partial\pq{\LLL_1(\theta),\dots,\LLL_M(\theta)}}{\partial\theta^T}\right|_{\theta=\hat\theta(\llambda)}.
\end{equation}
The local dependence of the losses on the scalarization weights can be expressed as a bilinear form evaluated on $\frac{\partial\LLL_i}{\partial\theta}$ and $\frac{\partial\LLL_j}{\partial\theta}$:
\begin{equation}
    \frac{\partial\LLL_i(\hat\theta(\llambda))}{\partial\lambda_j}
    =\left.\frac{\partial\LLL_i}{\partial\theta}\right|_{\theta=\hat\theta(\llambda)}\frac{\partial\hat\theta(\llambda)}{\partial\lambda_j}\\
    =-\left.\frac{\partial\LLL_i}{\partial\theta}\AA^{-1}\frac{\partial\LLL_j}{\partial\theta^T}\right|_{\theta=\hat\theta(\llambda)}.
    \label{eqn:derivative_inner}
\end{equation}
Because $\hat\theta$ is a local minimizer, $-\AA^{-1}$ is negative definite. In particular, any $\frac{\partial\LLL_i(\hat\theta(\llambda))}{\partial\lambda_i}$ is negative. This expresses the intuitive fact that if an infinitesimally higher weight is given to some prediction loss in optimization, the value of this loss at the optimum will be infinitesimally lower.

For concreteness, consider how the highest-length prediction loss $\LLL_M(\hat\theta(\llambda))$ changes when $\lambda_M$ is increased and the $\lambda_j$ ($j\neq i$) are decreased with rate proportional to $\lambda_j$, while $\sum\lambda_j$ is kept constant. That is, let $\bbeta=\pq{-\lambda_1,\dots,-\lambda_{i-1},\sum_{j\neq i}\lambda_j,-\lambda_{i+1},\dots,-\lambda_M}$. Then
\begin{equation}
    \frac{d\LLL_i(\hat\theta(\llambda+t\bbeta))}{dt}
    =\sum_j\frac{\partial\LLL_i}{\partial\lambda_j}\beta_j
    =-\frac{\partial\LLL_i}{\partial\theta}\AA^{-1}\sum_j\frac{\partial\LLL_j}{\partial\theta^T}\beta_j
    =-\frac{\partial\LLL_i}{\partial\theta}\AA^{-1}\frac{\partial\LLL_i}{\partial\theta^T}\sum_j\lambda_j\leq0,
\end{equation}
where the last two equalities follow from (\ref{eqn:derivative_inner}) and (\ref{eqn:derivative_alignment}), respectively, and the inequality holds because $\AA^{-1}$ is positive definite. So we have shown that, in nondegenerate cases, the $\LLL_M(\theta)$ term of the optimization criterion decreases under the locally optimal weights $\theta$ when $\lambda_M$ is infinitesimally increased in this way.

\paragraph{Log-linear mixture of predictors.}

Returning to coherence boosting, suppose that we aim to build out of the predictors $f_k(-;\theta\hat(\llambda))$ a new predictor $g$ that would have lower negative log-likelihood on prediction of a word given the maximum-length context:
\[
\EE_{w_1\dots w_{M+1}\in\DDD}\left[-\log g(w_{M+1}\mid w_1,\dots,w_M)\right]
<
\EE\left[-\log f_M(w_{M+1}\mid w_1,\dots,w_M;\hat\theta(\llambda))\right].
\]
As we just saw, using this predictor in place of $f_M$ achieves the same direction of movement in the prediction loss as optimizing with higher weight $\lambda_M$.

A na\"ive guess -- not a proper predictor, as its outputs do not sum to 1 -- would lightly perturb $f_M$ by log-linearly mixing small multiples of the $f_k$ weight weights $\beta_k$ summing to 0: 
\[
g_{\text{\rm na\"ive}}^{(t)}(w_1,\dots,w_M)=\exp\left(\log f_M(w_1,\dots,w_M;\hat\theta(\llambda))+t\sum_k\beta_k\log f_k(-,\hat\theta(\llambda))\right).
\]
Then, by linearity of expectation,
\begin{align}
    \left.\frac{d}{dt}\right|_{t=0}\EE\left[-\log g_{\text{\rm na\"ive}}^{(t)}(w_{M+1}\mid w_1,\dots,w_M)\right]
    &=\sum_k\beta_k\EE\left[-\log f_k(w_{M+1}\mid w_1,\dots,w_M;\hat\theta(\llambda))\right]\nonumber\\
    &=\sum_k\beta_k\LLL_k(\hat\theta(\llambda)).\label{eqn:derivative_naive}
\end{align}
This quantity is negative if, for example, $\LLL_M(\hat\theta(\llambda))$ is minimal among the $\LLL_k(\hat\theta(\llambda))$.

Reintroducing the normalization condition, we define a candidate function $g^{(t)}$ as the normalization of $g_{\text{\rm na\"ive}}^{(t)}$ over $w_{M+1}$ and compute, with the aid of (\ref{eqn:derivative_naive}) and using that the $g_k$ are normalized to simplify the derivative of $\log\sum\exp$:
\begin{align}
    &\left.\frac{d}{dt}\right|_{t=0}\EE\left[-\log g^{(t)}(w_{M+1}\mid w_1,\dots,w_M)\right]\nonumber\\
    &=\sum_k\beta_k\LLL_k(\hat\theta(\llambda)) + \left.\frac{d}{dt}\right|_{t=0}\EE\log\sum_wg_{\text{\rm na\"ive}}^{(t)}(w\mid w_1,\dots,w_M)\nonumber\\
    &=\sum_k\beta_k\LLL_k(\hat\theta(\llambda)) + \EE\sum_w\gq{\sum_k\beta_k\log f_k(w_1,\dots,w_M;\hat\theta(\llambda)), f_M(w_1,\dots,w_M;\hat\theta(\llambda))}\nonumber\\
    &=\sum_k\beta_k\LLL_k(\hat\theta(\llambda)) - \sum_k\beta_k\EE\bq{D_{\rm KL}\pq{f_M(w_1,\dots,w_M;\hat\theta(\llambda))\,\|\,f_k(w_1,\dots,w_M;\hat\theta(\llambda))}},
\end{align}
where the last line used that $\sum \beta_k=0$.

In practice, we are interested in sparse log-linear mixtures. Taking $\beta_M=1$, $\beta_k=-1$ for a single $k$, and all other $\beta_i=0$, we conclude that the boosted model proportional to $f_M^{1+t}f_k^{-t}$ is a better predictor than $f_M$ alone if the difference between prediction losses $\LLL_M$ and $\LLL_k$ is greater than the average KL divergence between the predictions $f_M$ and $f_k$.

\section{From coherence boosting to coherence tuning}
\label{sec:coherence_tuning} 

As mentioned in the main text, algorithms that modify the weights of a pretrained LM to increase effect of distant words, mimicking coherence boosting, are an interesting direction for future work. Here we propose an algorithm, \textbf{coherence tuning}, that achieves this without training on any specialized data. 

Initializing with the pretrained model $f(-|-;\theta)$, the algorithm iterates the following training steps to bring the LM closer to its coherence-boosted version $f_{\aalpha}$:
\begin{enumerate}[(1)]
\item Generate a sequence $w_1\dots w_n$ from the current model $f(-|-;\theta)$.
\item Compute all next-token distributions under the coherence-boosted version of the current model ($f_{\aalpha}(w_1\dots w_k;\theta)$) and under the current model without boosting ($f(w_1\dots w_k;\theta)$).
\item Gradient step on ${\rm KL}(f_{\aalpha}(w_1\dots w_k;\theta)\|f(w_1\dots w_k;\theta))$, where the first distribution $f_{\aalpha}$ is treated as constant. This step may be restricted only to $k$ near the end of the sequence.
\end{enumerate}

\begin{figure}[t]
\centering
\includegraphics[width=0.8\textwidth]{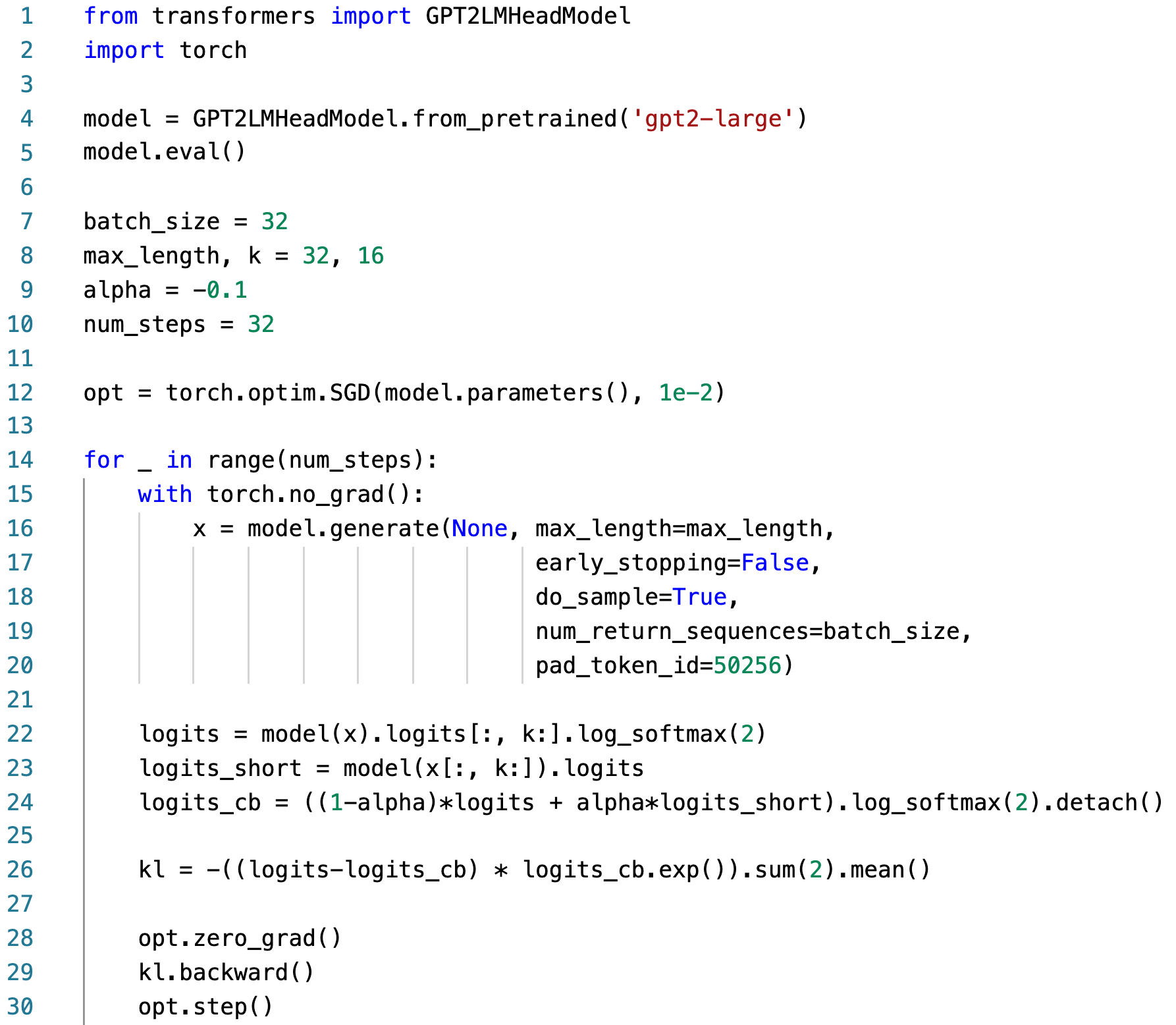}
\caption{Coherence tuning in PyTorch.}
\label{fig:tuning-code}
\end{figure}

We provide a batched implementation in Fig.~\ref{fig:tuning-code} in lieu of pseudocode. This coherence tuning code, which performs 32 gradient steps on batches of 32 sequences of length 32, runs in a few minutes on modern hardware, amortizing the overhead cost of coherence boosting while achieving comparable results on the WebText article completion task (second-to-last row of Table~\ref{tab:owt-gen}). 

\section{GPT-2 summarization experiments}
\label{sec:summarization}

In \S\ref{sec:experiments_generation} of the main text, we applied coherence boosting to generic text and dialogue response generation. Another interesting task that also requires long-range coherence is text summarization, in which the model is often expected to attend to the first few sentences to summarize a long article. Thus, we provide preliminary experiments for zero-shot abstractive summarization by applying our proposed method to GPT-2 models.

\paragraph{Experiment details.}
We take the two most popular summarization datasets, CNN/DM~\cite{see-etal-2017-get} and XSum~\cite{narayan2018don}, where both contain recent articles and the summaries for the latter are more abstractive than the former. Following standard design~\cite{radford2019language}, we append the  tokens ``\textsc{TL;DR:}'' at the end of each article to induce summarization behavior of GPT models. We leverage the GPT-2 XL model and let it continue generating 100 tokens with greedy decoding. We take the first three sentences for CNN/DM articles and the first two sentences for XSum articles as their summaries. We use the preprocessed data and metric calculation from \citet{zhong2020extractive} and report the standard ROUGE scores in Table~\ref{tab:gpt_abs_sum}.

To apply our proposed coherence boosting method, similarly to the method used for dialogue response generation, we define the short context as the newly generated text after the ``\textsc{TL;DR:}'' tokens. That is, at any time step during the summarization, the long context is the full article with the so-far generated summary, and the short context is only the generated summary. 

\begin{table}[h]
\centering
\begin{tabular}{lcccccc}
\toprule
                 & \multicolumn{3}{c}{CNN/DM}                         & \multicolumn{3}{c}{XSum}                           \\
                 \cmidrule(lr){2-4}\cmidrule(lr){5-7}
                 & ROUGE-1         & ROUGE-2        & ROUGE-L         & ROUGE-1         & ROUGE-2        & ROUGE-L         \\ \midrule
GPT-2 XL           & 26.671          & 7.792          & 23.926          & 21.346          & \textbf{4.360} & 16.880          \\ \midrule
CB, $\alpha=-0.1$ & 28.027          & 8.658          & 25.179          & \textbf{21.580} & {\ul 4.265}    & {\ul 17.025}    \\
CB, $\alpha=-0.2$ & 28.995          & 9.293          & 26.066          & {\ul 21.571}    & 4.200          & \textbf{17.026} \\
CB, $\alpha=-0.3$ & 29.502          & 9.528          & 26.442          & 21.405          & 4.045          & 16.848          \\
CB, $\alpha=-0.4$ & 29.772          & \textbf{9.663} & {\ul 26.644}    & 21.150          & 3.876          & 16.613          \\
CB, $\alpha=-0.5$ & \textbf{29.872} & {\ul 9.625}    & \textbf{26.658} & 20.773          & 3.703          & 16.288          \\
CB, $\alpha=-0.6$ & {\ul 29.827}    & 9.500          & 26.524          & 20.379          & 3.525          & 16.010          \\
CB, $\alpha=-0.7$ & 29.742          & 9.392          & 26.399          & 20.063          & 3.437          & 15.776          \\
CB, $\alpha=-0.8$ & 29.703          & 9.304          & 26.242          & 19.661          & 3.305          & 15.460          \\
CB, $\alpha=-0.9$ & 29.481          & 9.109          & 25.990          & 19.387          & 3.195          & 15.256          \\ \bottomrule
\end{tabular}
\caption{Abstractive summarization performance with the GPT-2 XL model. The best performance is bolded and the second-best is underlined.}
\label{tab:gpt_abs_sum}
\end{table}

\paragraph{Results.}
As we can see from Table~\ref{tab:gpt_abs_sum}, our proposed CB method improves most of the metrics on both datasets. On the CNN/DM dataset, CB yields improvements of up to $\sim$3 ROUGE points. We believe such a significant improvement is due to the article structure of the CNN/DM dataset. Specifically, the first three sentences in CNN/DM articles can provide pretty good summaries for a large portion of articles and have been considered as a very strong baseline for summarization models~\cite{zhong2020extractive}. This long-range information in the first few sentences is hard to capture for GPT models, and this difficulty can be alleviated by coherence boosting. 

Such structure might be less pronounced in the XSum dataset, thus leading to a smaller improvement from coherence boosting. The difference between the datasets is also reflected in the best value of $\alpha$. The optimal $\alpha$ for the CNN/DM dataset is around $-$0.5, while the best $\alpha$ for XSum is around $-$0.2, which indicates that boosting has less effect on XSum. 

Finally, we note that the performance with GPT models presented here is far lower than that of state-of-the-art summarization-specific models, such as finetuned BART models \citep{lewis2020bart, zhang2020pegasus}. We leave the use of coherence boosting with such encoder-decoder models to future work. 


\newpage

\section{Example WebText completion}

An example of human, top-$p$, and coherence boosting outputs is shown in Table~\ref{tab:owt-examples}. All outputs for all boosting schemes are included in the code repository.

\begin{table*}[h]
\newcommand{\nonsense}[1]{#1}
\newcommand{\relevant}[1]{#1}
\small
\resizebox{\linewidth}{!}{
\begin{tabular}{p{0.1\textwidth}p{1\textwidth}}
\toprule
{\large Prompt}
&{\large University of Alberta doctoral candidate Scott Persons measured the limb length for 53 different species of carnivorous dinosaurs to create an equation to score how much each was built for speed.}
\\\midrule
{\large Human} 
&{ \large A new study published out of the University of
\relevant{Alberta} has found that carnivorous
\relevant{dinosaurs} evolved to become faster and faster over time — and the proof is in the leg length.
As part of his
\relevant{doctoral} research, lead author
\relevant{Scott}
\relevant{Persons} travelled the world measuring the limb
\relevant{lengths} for
\relevant{53} different species of
\relevant{carnivorous} bipedal
\relevant{dinosaurs}, including the Velociraptor, Allosaurus and Tyrannosaurus rex.
In particular,
\relevant{Persons} looked at the length of the dinosaurs' legs below the knee. As a rule, he said, the
\relevant{longer} the lower leg is in comparison to the upper leg, the
\relevant{faster} the
\relevant{animal} is.
Modern-day cheetahs have proportionately longer legs than hyenas. That relationship is mirrored in the animals' speeds, Persons said.
"That's true for modern carnivores, and must have been true for
\relevant{dinosaurs},"
\relevant{Persons} said in a statement released Wednesday. }
 \\\midrule
{\large Nucleus (\mbox{$p=0.95$})}
&{\large "In a lot of areas of the world, dinosaurs at the time didn't have legs, they had weapons," said
\relevant{Persons}.
So he used his scientific skill, \nonsense{titled "Toill's Law of Tragedy,"} to \nonsense{give the dinosaurs an on-the-ground estimate} of what their internal organs weighed.
For example, wood-eating reptiles, such as the dinosaur Lufengosaurus and the \nonsense{ceratopsian Tyrannosaurus rex}, would have
\relevant{weighed} up to 52 kg (115 lbs)
\relevant{each}. And as many as 50 \nonsense{plates} should have been available to hunt,
\relevant{Persons} said.
But what's important is that a
\relevant{dinosaur's} pack or herd probably wouldn't have been able to feed \nonsense{all three} in one sitting.
D\relevant{inosaurs} were extremely \nonsense{resilient, as very few of them survived from one kill to the next} and later on would start a \nonsense{``phased trip'' back to where they'd been
\relevant{killed}},
\relevant{Persons} said.}
 \\\midrule
{\large +$\,\,\,$boost \mbox{($k=64$,} \mbox{$\alpha_k=-0.1$)}}
&{\large "In a lot of areas of the world, dinosaurs at the time didn't have legs, they had weapons," said
\relevant{Persons}.
So he used his undergraduate course to work on a formula to calculate a specific height for each carnivore based on the evolutionary requirements of different dinosaurs and when it would have been optimal for meat eating.
He then moved that
\relevant{formula} into his
\relevant{doctoral} thesis project, which is looking at limb lengths
\relevant{for} medium-sized, fearsome
\relevant{carnivores}.
"His thesis is about how to measure the \nonsense{shape of different places, plant diversity} and evolutionary biology and kind of how we see dinosaurs and the relationships between species," said Susanne Dufoot, an associate professor of paleontology at McMaster University who recently returned from attending
\relevant{Persons}' research.
"It's interesting because he's basically done the legwork, developed this model that can give us information about \nonsense{plant species}."
\nonsense{`He was an amazing creature'}}
\\\bottomrule
\end{tabular}
}
\caption{Completions of an article: written by a human (original WebText) and sampled from GPT-2 Large with top-$p$ sampling, with and without coherence boosting. While top-$p$ sampling produces text that is coherent at first glance -- it is free of repetition and nonce words -- the topic of the article meanders from limb length to internal organs and killing, and nonsensical comments appear (`Toill's Law of Tragedy', herbivorous ceratopsian T-Rex, etc.). The output with coherence boosting is largely free of these issues, maintaining focus on limb length and diet.}
\label{tab:owt-examples}
\end{table*}

\newpage

\section{Prompt formats for multiple-choice tasks}
\label{sec:prompt_formats}

\newcommand\blueuline{\bgroup\markoverwith
{\textcolor{blue}{\rule[-0.75ex]{2.8pt}{0.7pt}}}\ULon}

\newcommand\reduline{\bgroup\markoverwith
{\textcolor{red}{\rule[-1.2ex]{0.1pt}{0.7pt}}}\ULon}

\newcommand\phuline{\bgroup\markoverwith
{\textcolor{white}{\rule[-1.2ex]{0.1pt}{0.7pt}}}\ULon}

\def\word{\textcolor{gray}{\rm w}}
\def\full{\textcolor{blue}{\rm full}}
\def\short{\textcolor{red}{\rm short}}
\newcommand{\mybox}[1]{#1}

\begin{table*}[h]
\centering
\resizebox{\linewidth}{!}{

\begin{tabular}{ll}
\toprule
Task        & Prompt format \\ \midrule
Story Cloze & 
\blueuline{[Context]\reduline{ }} [Completion]
\\ \midrule
HellaSwag & 
\blueuline{[Context] \reduline{he/she/they/...}} [Completion]
\\ \midrule
COPA & 
\blueuline{[Premise] \reduline{because/so}} [Hypothesis]
\\ \midrule
CommonsenseQA & 
\blueuline{[Question] \reduline{the answer is:}} [Answer]
\\ \midrule
OpenBookQA & 
\blueuline{[Question] \reduline{the answer is:}} [Answer]
\\ \midrule
ARC Easy & 
\blueuline{Question: [Question] \reduline{Answer:}} [Answer]
\\ \midrule
ARC Challenge & 
\blueuline{Question: [Question] \reduline{Answer:}} [Answer]
\\ \midrule
PIQA & 
\blueuline{Question: [Question] \reduline{Answer:}} [Answer]
\\ \midrule
SST-2 & 
\blueuline{[Context] \reduline{This quote has a tone that is:}} [Label] \\ \midrule
SST-5 & 
\blueuline{[Context] \reduline{This quote has a tone that is:}} [Label] \\ \midrule
AGNews & 
\blueuline{Title: [Title]\ Summary: [Context] \reduline{\\Topic:}} [Label] \\ \midrule
TREC & 
\blueuline{[Question] \reduline{The answer to this question will be}} [Label] \\ \midrule
BoolQ & 
\blueuline{[Passage]\textbackslash{n} Question: [Hypothesis] \reduline{True or False? Answer:}} [Label] \\ \midrule
RTE & 
\blueuline{[Premise]\textbackslash{n}\reduline{ question: [Hypothesis] true or false?\textbackslash{n} answer:}} [Label] \\ \midrule
CB & 
\blueuline{Given question: [Premise]\reduline{ Is [Hypothesis] true, false or neither?\textbackslash{n} The answer is:}} [Label] \\ 
\bottomrule
\end{tabular}
}
\caption{Prompt formats used in our experiments. The full context is underlined in blue; the premise-free context is also underlined in red. We mainly draw inspiration from \cite{brown2020language, holtzman2021surface, zhao2021calibrate} to make our prompts more natural to facilitate boosting the coherence of the completion.}
\label{tab:prompt_format}

\end{table*}

\newpage
\section{Additional results}

\begin{table*}[h]
\centering
\resizebox{1\linewidth}{!}{
\begin{tabular}{@{}l|cccc|cccc|cccc|cccc@{}}
\toprule
            & \multicolumn{4}{c|}{GPT-3 Small}                                   & \multicolumn{4}{c|}{GPT-3 Medium}                                  & \multicolumn{4}{c|}{GPT-3 Large}                                   & \multicolumn{4}{c}{GPT-3 XL}                                      \\
            & $f_{\max}$    & $\alpha=1$    & $\alpha=\alpha^*$ & $\alpha^*$    & $f_{\max}$    & $\alpha=1$    & $\alpha=\alpha^*$ & $\alpha^*$    & $f_{\max}$    & $\alpha=1$    & $\alpha=\alpha^*$ & $\alpha^*$    & $f_{\max}$    & $\alpha=1$    & $\alpha=\alpha^*$ & $\alpha^*$    \\ \midrule
Story Cloze & 66.0          & 70.9          & \textbf{74.5}     & \textit{-0.8} & 70.1          & 76.3          & \textbf{78.0}     & \textit{-0.8} & 74.2          & \textbf{82.9} & 80.8              & \textit{-0.7} & 79.3          & 82.9          & \textbf{86.9}     & \textit{-0.6} \\
HellaSwag   & 35.7          & 38.9          & \textbf{42.0}     & \textit{-0.9} & 42.8          & 46.8          & \textbf{51.3}     & \textit{-0.8} & 50.5          & 55.1          & \textbf{62.2}     & \textit{-0.8} & 59.2          & 62.7          & \textbf{72.3}     & \textit{-0.8} \\
COPA        & 73.0          & 71.0          & \textbf{75.0}     & \textit{-0.6} & \textbf{85.0} & 79.0          & 83.0              & \textit{-0.7} & \textbf{84.0} & 83.0          & \textbf{84.0}     & \textit{-0.6} & 93.0          & 87.0          & \textbf{94.0}     & \textit{-0.5} \\ \midrule
CsQA        & 34.6          & 46.4          & \textbf{48.0}     & \textit{-0.7} & 42.4          & 51.4          & \textbf{53.0}     & \textit{-0.7} & 50.0          & 57.5          & \textbf{60.4}     & \textit{-0.7} & 61.1          & 68.0          & \textbf{70.4}     & \textit{-0.7} \\
OBQA        & 16.0          & 39.8          & \textbf{46.6}     & \textit{-2.2} & 16.4          & 41.8          & \textbf{48.8}     & \textit{-1.4} & 20.8          & 45.4          & \textbf{47.8}     & \textit{-1.6} & 28.0          & 52.2          & \textbf{52.6}     & \textit{-1.1} \\
ARC-E       & 51.3          & 48.1          & \textbf{56.0}     & \textit{-0.5} & 59.8          & 54.8          & \textbf{63.3}     & \textit{-0.4} & 68.4          & 60.3          & \textbf{70.7}     & \textit{-0.5} & 76.2          & 69.2          & \textbf{78.3}     & \textit{-0.4} \\
ARC-C       & 22.6          & 30.8          & \textbf{31.1}     & \textit{-1.4} & 27.5          & 35.3          & \textbf{35.5}     & \textit{-1.2} & 33.9          & \textbf{41.8} & \textbf{41.8}     & \textit{-0.9} & 43.9          & \textbf{50.6} & 49.2              & \textit{-1.1} \\
PIQA        & 69.0          & 57.5          & \textbf{69.6}     & \textit{-0.4} & 74.4          & 60.4          & \textbf{74.7}     & \textit{-0.4} & 76.3          & 64.2          & \textbf{77.7}     & \textit{-0.4} & \textbf{79.3} & 66.3          & 78.9              & \textit{-0.6} \\ \midrule
SST-2       & 70.6          & 79.8          & \textbf{84.6}     & \textit{-2.3} & 69.5          & 75.2          & \textbf{88.0}     & \textit{-4.8} & 66.8          & 65.2          & \textbf{70.0}     & \textit{2.0}  & 86.2          & 88.1          & \textbf{89.8}     & \textit{-0.5} \\
SST-5       & \textbf{26.7} & 26.6          & 26.1              & \textit{-1.1} & 29.3          & \textbf{30.7} & 30.0              & \textit{-1.2} & 28.1          & \textbf{33.2} & 30.1              & \textit{-0.8} & 31.2          & 34.8          & \textbf{38.5}     & \textit{-1.4} \\
AGNews      & 67.1          & 69.2          & \textbf{69.5}     & \textit{-1.2} & 63.3          & 64.8          & \textbf{65.4}     & \textit{-2.0} & 69.2          & 65.7          & \textbf{69.5}     & \textit{-0.3} & 71.7          & 71.7          & \textbf{71.8}     & \textit{0.2}  \\
TREC        & 28.8          & 57.2          & \textbf{57.4}     & \textit{-1.0} & 30.2          & 62.6          & \textbf{63.6}     & \textit{-0.8} & 35.2          & 28.8          & \textbf{37.2}     & \textit{-0.3} & 52.4          & 47.0          & \textbf{56.0}     & \textit{-0.6} \\
BoolQ       & 60.7          & \textbf{62.4} & 62.2              & \textit{-1.4} & 61.6          & 63.4          & \textbf{63.5}     & \textit{-0.9} & 64.2          & 65.6          & \textbf{68.1}     & \textit{-4.5} & 71.6          & \textbf{73.7} & 72.7              & \textit{-0.4} \\ \midrule
RTE         & 49.8          & \textbf{51.3} & \textbf{51.3}     & \textit{-3.6} & \textbf{54.5} & 50.5          & 49.1              & \textit{-1.2} & 53.8          & \textbf{55.6} & 55.2              & \textit{-1.4} & 56.0          & 57.4          & \textbf{60.3}     & \textit{-0.6} \\
CB          & \textbf{33.9} & 19.6          & 21.4              & \textit{-0.7} & 8.9           & 25.0          & \textbf{39.3}     & \textit{-1.9} & \textbf{32.1} & 28.6          & \textbf{32.1}     & \textit{-0.2} & 5.4           & 25.0          & \textbf{28.6}     & \textit{-1.9} \\ \midrule
average & 47.1 & 51.3 & \textbf{54.4} & \textit{$-$1.3} & 49.0 & 54.5 & \textbf{59.1} & \textit{$-$1.3} & 53.8 & 55.5 & \textbf{59.2} & \textit{$-$0.8} & 59.6 & 62.4 & \textbf{66.7} & \textit{$-$0.7} \\
\bottomrule
\end{tabular}

}
\caption{Accuracy (\%) of GPT-3 models on all multiple-choice tasks, in the same format as Table~\ref{tab:multichoice-main}.}
\label{tab:multichoice-gpt3}
\end{table*}

\begin{table*}[h]
\centering
\resizebox{1\linewidth}{!}{
\begin{tabular}{@{}l|cccc|cccc|cccc|cccc@{}}
\toprule
            & \multicolumn{4}{c|}{GPT-2 Small}                            & \multicolumn{4}{c|}{GPT-2 Medium}                           & \multicolumn{4}{c|}{GPT-2 Large}                               & \multicolumn{4}{c}{GPT-2 XL}                                  \\
            & $f_{\max}$ & $\alpha=-1$   & Ours          & $\alpha^*$    & $f_{\max}$ & $\alpha=-1$   & Ours          & $\alpha^*$    & $f_{\max}$    & $\alpha=-1$   & Ours          & $\alpha^*$    & $f_{\max}$    & $\alpha=-1$   & Ours          & $\alpha^*$    \\ \midrule
Story Cloze & 59.9       & \textbf{64.8} & 64.2          & \textit{-1.0} & 63.0       & 68.5          & \textbf{70.4} & \textit{-0.7} & 66.0          & 72.0          & \textbf{74.4} & \textit{-0.8} & 67.6          & 75.1          & \textbf{76.8} & \textit{-0.7} \\
HellaSwag   & 28.9       & 31.0          & \textbf{31.8} & \textit{-0.9} & 33.4       & 36.6          & \textbf{38.1} & \textit{-0.9} & 36.6          & 39.5          & \textbf{43.0} & \textit{-0.8} & 40.0          & 42.6          & \textbf{47.7} & \textit{-0.8} \\
COPA        & 62.0       & 56.0          & \textbf{64.0} & \textit{-0.7} & 69.0       & 69.0          & \textbf{72.0} & \textit{-0.6} & \textbf{69.0} & 60.0          & \textbf{69.0} & \textit{-0.6} & 73.0          & 70.0          & \textbf{77.0} & \textit{-0.4} \\ \midrule
CsQA        & 29.5       & 42.3          & \textbf{43.2} & \textit{-0.8} & 31.3       & 44.6          & \textbf{45.3} & \textit{-0.8} & 35.7          & 47.3          & \textbf{50.0} & \textit{-0.8} & 37.8          & 50.5          & \textbf{52.9} & \textit{-0.8} \\
OBQA        & 11.2       & 30.6          & \textbf{40.8} & \textit{-1.6} & 15.6       & 34.8          & \textbf{43.8} & \textit{-2.1} & 13.6          & 34.4          & \textbf{44.2} & \textit{-1.8} & 15.6          & 38.4          & \textbf{47.0} & \textit{-1.9} \\
ARC-E       & 43.8       & 42.1          & \textbf{46.0} & \textit{-0.3} & 49.1       & 44.5          & \textbf{51.3} & \textit{-0.6} & 53.2          & 46.5          & \textbf{56.2} & \textit{-0.5} & 58.3          & 51.4          & \textbf{60.3} & \textit{-0.4} \\
ARC-C       & 19.0       & 26.1          & \textbf{29.1} & \textit{-4.2} & 21.5       & \textbf{27.3} & 27.0          & \textit{-1.0} & 21.7          & 28.3          & \textbf{29.1} & \textit{-2.8} & 25.0          & 33.5          & \textbf{34.4} & \textit{-1.1} \\
PIQA        & 62.9       & 57.5          & \textbf{63.4} & \textit{-0.6} & 67.6       & 56.1          & \textbf{68.1} & \textit{-0.5} & \textbf{70.3} & 60.0          & 70.1          & \textit{-0.4} & 70.8          & 60.4          & \textbf{71.5} & \textit{-0.4} \\ \midrule
SST-2       & 65.7       & 74.7          & \textbf{82.3} & \textit{-2.2} & 72.6       & 83.5          & \textbf{88.2} & \textit{-2.0} & 77.2          & 87.6          & \textbf{88.0} & \textit{-1.2} & 86.4          & 84.5          & \textbf{86.9} & \textit{-0.1} \\
SST-5       & 25.9       & \textbf{30.9} & \textbf{30.9} & \textit{-1.2} & 20.5       & 33.3          & \textbf{35.2} & \textit{-1.1} & 29.1          & 31.8          & \textbf{35.2} & \textit{-1.4} & 28.7          & \textbf{38.7} & 36.9          & \textit{-1.7} \\
AGNews      & 58.6       & 60.8          & \textbf{62.2} & \textit{-0.6} & 64.6       & \textbf{66.5} & 66.3          & \textit{-0.7} & 62.6          & 62.1          & \textbf{63.8} & \textit{-0.4} & 67.2          & 67.4          & \textbf{68.3} & \textit{-0.4} \\
TREC        & 23.4       & 29.6          & \textbf{32.2} & \textit{-0.8} & 27.4       & 17.6          & \textbf{36.0} & \textit{-0.4} & 22.6          & \textbf{45.4} & 44.2          & \textit{-1.2} & 23.4          & 27.4          & \textbf{40.0} & \textit{-0.8} \\
BoolQ       & 49.4       & 58.1          & \textbf{62.1} & \textit{-3.0} & 56.6       & \textbf{61.8} & \textbf{61.8} & \textit{-0.9} & 61.2          & \textbf{62.3} & 62.2          & \textit{-1.8} & 62.1          & \textbf{63.5} & 63.2          & \textit{-0.6} \\ \midrule
RTE         & 51.3       & 49.8          & \textbf{53.4} & \textit{-0.3} & 53.1       & 50.9          & \textbf{53.8} & \textit{-0.2} & \textbf{53.1} & 46.6          & 50.2          & \textit{-1.2} & \textbf{49.1} & 48.7          & \textbf{49.1} & \textit{0.9}  \\
CB          & 12.5       & 23.2          & \textbf{48.2} & \textit{-2.4} & 8.9        & 37.5          & \textbf{55.4} & \textit{-2.5} & 8.9           & 32.1          & \textbf{53.6} & \textit{-2.5} & 30.4          & 51.8          & \textbf{66.1} & \textit{-1.9}
\\ \midrule
average & 40.3&45.2&\textbf{50.3}&\textit{$-$1.4}&43.6&48.8&\textbf{54.2}&\textit{$-$1.0}&45.4&50.4&\textbf{55.5}&\textit{$-$1.2}&49.0&53.6&\textbf{58.5}&\textit{$-$0.7}
\\\bottomrule
\end{tabular}
}
\caption{Accuracy (\%) of GPT-2 models on all multiple-choice tasks, in the same format as Table~\ref{tab:multichoice-main}.}
\label{tab:multichoice-gpt2}
\end{table*}

\begin{table*}[h]
\centering
\resizebox{0.8\linewidth}{!}{
\begin{tabular}{@{}l|ccc|cc|cc|ccc@{}}
\toprule
            & \multicolumn{3}{c|}{GPT-3 Small}      & \multicolumn{2}{c|}{GPT-3 Medium} & \multicolumn{2}{c|}{GPT-3 Large} & \multicolumn{3}{c}{GPT-3 XL}         \\
            & PMI           & CC   & Ours          & PMI             & Ours           & PMI            & Ours           & PMI           & CC   & Ours          \\ \midrule
Story Cloze & 73.1          & -    & \textbf{74.5} & 76.8            & \textbf{78.0}  & 79.9           & \textbf{80.8}  & 84.0          & -    & \textbf{86.9} \\
HellaSwag   & 34.2          & -    & \textbf{42.0} & 40.0            & \textbf{51.3}  & 45.8           & \textbf{62.2}  & 53.5          & -    & \textbf{72.3} \\
COPA        & 74.4          & -    & \textbf{75.0} & 77.0            & \textbf{83.0}  & \textbf{84.2}  & 84.0           & 89.2          & -    & \textbf{94.0} \\ \midrule
CsQA        & 44.7          & -    & \textbf{48.0} & 50.3            & \textbf{53.0}  & 58.5           & \textbf{60.4}  & 66.7          & -    & \textbf{70.4} \\
OBQA        & 42.8          & -    & \textbf{46.6} & 48.0            & \textbf{48.8}  & \textbf{50.4}  & 47.8           & \textbf{58.0} & -    & 52.6          \\
ARC-E       & 44.7          & -    & \textbf{56.0} & 51.5            & \textbf{63.3}  & 57.7           & \textbf{70.7}  & 63.3          & -    & \textbf{78.3} \\
ARC-C       & 30.5          & -    & \textbf{31.1} & 33.0            & \textbf{35.5}  & 38.5           & \textbf{41.8}  & 45.5          & -    & \textbf{49.2} \\ \midrule
SST-2       & 72.3          & 71.4 & \textbf{84.6} & 80.0            & \textbf{88.0}  & \textbf{81.0}  & 70.0           & 71.4          & 75.8 & \textbf{89.8} \\
SST-5       & 23.5          & -    & \textbf{26.1} & \textbf{32.0}   & 30.0           & 19.1           & \textbf{30.1}  & 29.6          & -    & \textbf{38.5} \\
AGNews      & 67.9          & 63.2 & \textbf{69.5} & 57.4            & \textbf{65.4}  & \textbf{70.3}  & 69.5           & \textbf{74.7} & 73.9 & 71.8          \\
TREC        & 57.2          & 38.8 & \textbf{57.4} & 61.6            & \textbf{63.6}  & 32.4           & \textbf{37.2}  & \textbf{58.4} & 57.4 & 56.0          \\
BoolQ       & 53.5          & -    & \textbf{62.2} & 61.0            & \textbf{63.5}  & 60.3           & \textbf{68.1}  & 64.0          & -    & \textbf{72.7} \\ \midrule
RTE         & \textbf{51.6} & 49.5 & 51.3          & 48.7            & \textbf{49.1}  & 54.9           & \textbf{55.2}  & \textbf{64.3} & 57.8 & 60.3          \\
CB          & \textbf{57.1} & 50.0 & 21.4          & 39.3            & 39.3           & \textbf{50.0}  & 32.1           & \textbf{50.0} & 48.2 & 28.6       \\ \bottomrule  
\end{tabular}
}
\caption{Performance comparison with other inference methods on GPT-3 models. PMI~\cite{holtzman2021surface} is an unconditional probability normalization method, CC~\cite{zhao2021calibrate} is the contextual calibration method. We compare them in the zero-shot setting.}
\label{tab:multichoice-pmi-gpt3}
\end{table*}

\begin{table*}[h]
\centering
\resizebox{0.8\linewidth}{!}{
\begin{tabular}{@{}l|cc|cc|ccc|ccc@{}}
\toprule
            & \multicolumn{2}{c|}{GPT-2 Small} & \multicolumn{2}{c|}{GPT-2 Medium} & \multicolumn{3}{c|}{GPT-2 Large}         & \multicolumn{3}{c}{GPT-2 XL}         \\
            & PMI            & Ours           & PMI             & Ours           & PMI           & Channel & Ours          & PMI           & CC   & Ours          \\ \midrule
Story Cloze & \textbf{67.0}  & 64.2           & \textbf{71.6}   & 70.4           & 73.4          & -       & \textbf{74.4} & 76.3          & -    & \textbf{76.8} \\
HellaSwag   & 29.1           & \textbf{31.8}  & 32.8            & \textbf{38.1}  & 35.1          & -       & \textbf{43.0} & 37.8          & -    & \textbf{47.7} \\
COPA        & 62.8           & \textbf{64.0}  & 70.0            & \textbf{72.0}  & \textbf{69.4} & -       & 69.0          & 71.6          & -    & \textbf{77.0} \\ \midrule
CsQA        & 36.4           & \textbf{43.2}  & 41.8            & \textbf{45.3}  & 44.5          & -       & \textbf{50.0} & 47.8          & -    & \textbf{52.9} \\
OBQA        & 32.4           & \textbf{40.8}  & 38.6            & \textbf{43.8}  & 43.2          & -       & \textbf{44.2} & 46.0          & -    & \textbf{47.0} \\
ARC-E       & 39.3           & \textbf{46.0}  & 42.4            & \textbf{51.3}  & 47.0          & -       & \textbf{56.2} & 49.9          & -    & \textbf{60.3} \\
ARC-C       & 28.2           & \textbf{29.1}  & \textbf{28.6}   & 27.0           & \textbf{31.6} & -       & 29.1          & 33.8          & -    & \textbf{34.4} \\ \midrule
SST-2       & 67.1           & \textbf{82.3}  & 86.2            & \textbf{88.2}  & 85.6          & 77.1    & \textbf{88.0} & \textbf{87.5} & 82.0 & 86.9          \\
SST-5       & 30.0           & \textbf{30.9}  & \textbf{39.3}   & 35.2           & 22.0          & 29.2    & \textbf{35.2} & \textbf{40.8} & -    & 36.9          \\
AGNews      & \textbf{63.0}  & 62.2           & 64.4            & \textbf{66.3}  & \textbf{64.1} & 61.8    & 63.8          & 65.4          & 60.0 & \textbf{68.3} \\
TREC        & \textbf{36.4}  & 32.2           & 21.6            & \textbf{36.0}  & 44.0          & 30.5    & \textbf{44.2} & 32.8          & 37.3 & \textbf{40.0} \\
BoolQ       & 51.1           & \textbf{62.1}  & 49.7            & \textbf{61.8}  & 46.7          & -       & \textbf{62.2} & 49.5          & -    & \textbf{63.2} \\ \midrule
RTE         & 49.8           & \textbf{53.4}  & \textbf{54.9}   & 53.8           & \textbf{54.2} & -       & 50.2          & \textbf{53.4} & 48.5 & 49.1          \\
CB          & \textbf{50.0}  & 48.2           & 50.0            & \textbf{55.4}  & 50.0          & -       & \textbf{53.6} & 50.0          & 17.9 & \textbf{66.1} \\
\bottomrule
\end{tabular}
}
\caption{Performance comparison with other inference methods on GPT-2 models. PMI~\cite{holtzman2021surface} is an unconditional probability normalization method, CC~\cite{zhao2021calibrate} is the contextual calibration method and Channel~\cite{min2021noisy} uses an inverted-LM scoring approach that computes the conditional probability of the input given the label. We compare them in the zero-shot setting.}
\label{tab:multichoice-pmi-gpt2}
\end{table*}

\begin{figure*}[h]
\centering
    \begin{subfigure}[t]{.48\textwidth}
        \centering
        \includegraphics[width=.9\textwidth]{figures/tasks/sweep_alpha_accuracy_storycloze.pdf}
    \end{subfigure}
    \hfill
    \begin{subfigure}[t]{.48\textwidth}
        \centering
        \includegraphics[width=.9\textwidth]{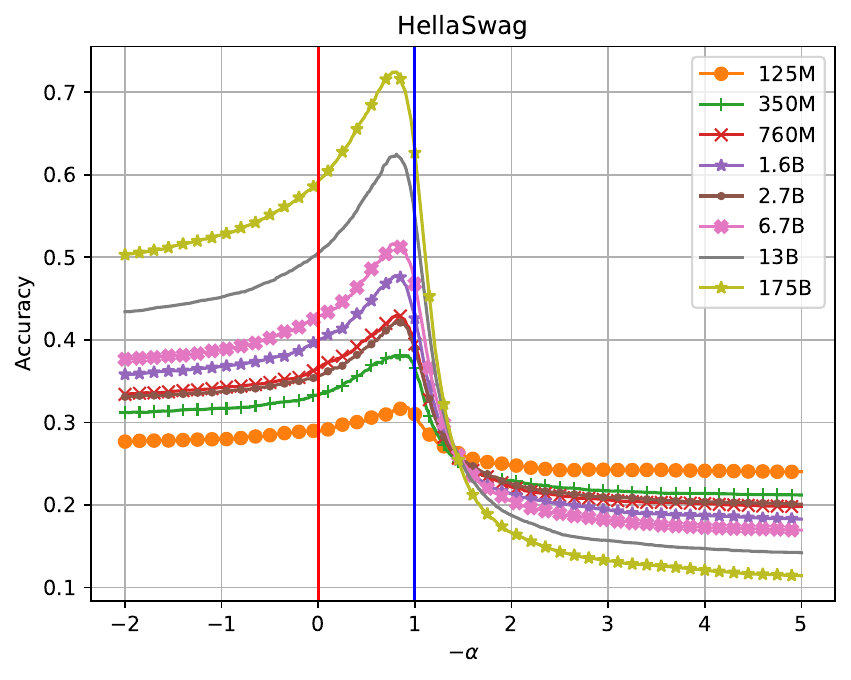}
    \end{subfigure}
    \vskip\baselineskip
    \begin{subfigure}[t]{.48\textwidth}
        \centering
        \includegraphics[width=.9\textwidth]{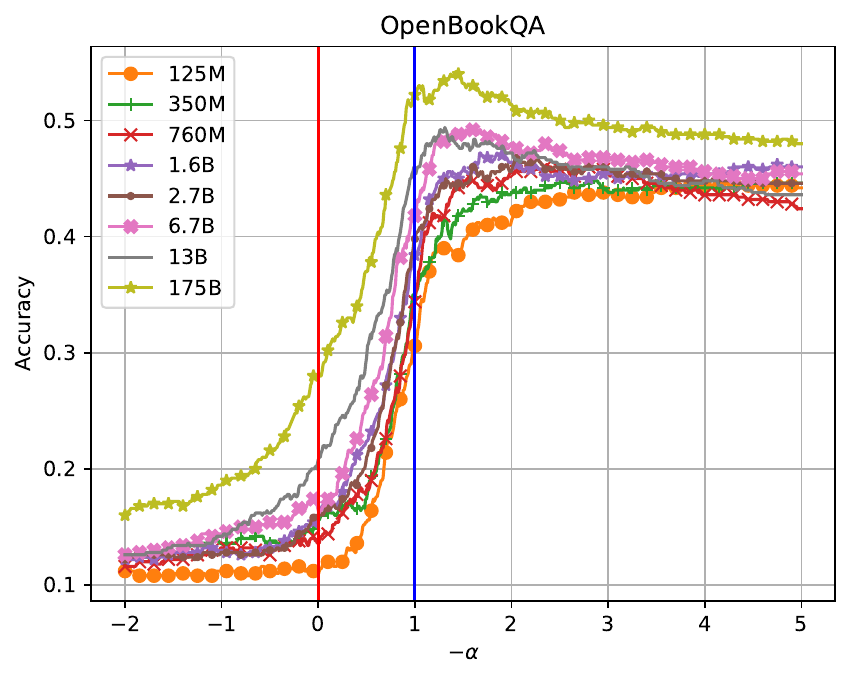}
    \end{subfigure}
    \hfill
    \begin{subfigure}[t]{.48\textwidth}
        \centering
        \includegraphics[width=.9\textwidth]{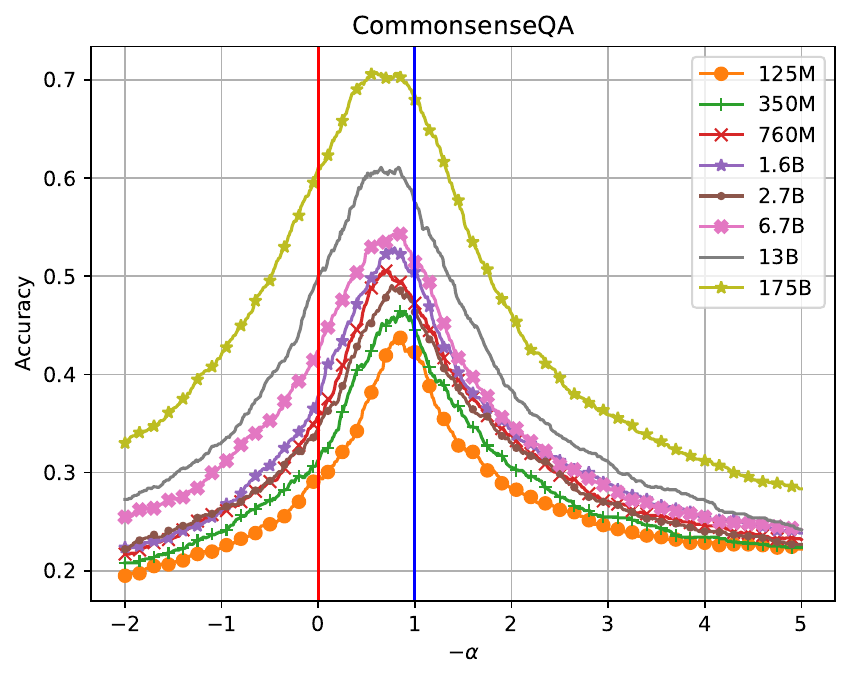}
    \end{subfigure}

    \vskip\baselineskip
    \begin{subfigure}[t]{.48\textwidth}
        \centering
        \includegraphics[width=.9\textwidth]{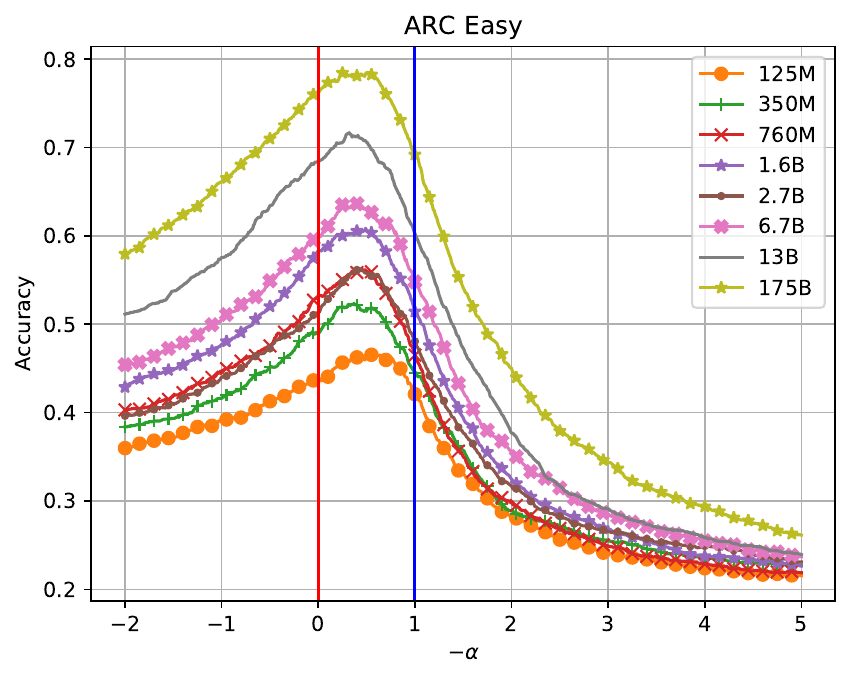}
    \end{subfigure}
    \hfill
    \begin{subfigure}[t]{.48\textwidth}
        \centering
        \includegraphics[width=.9\textwidth]{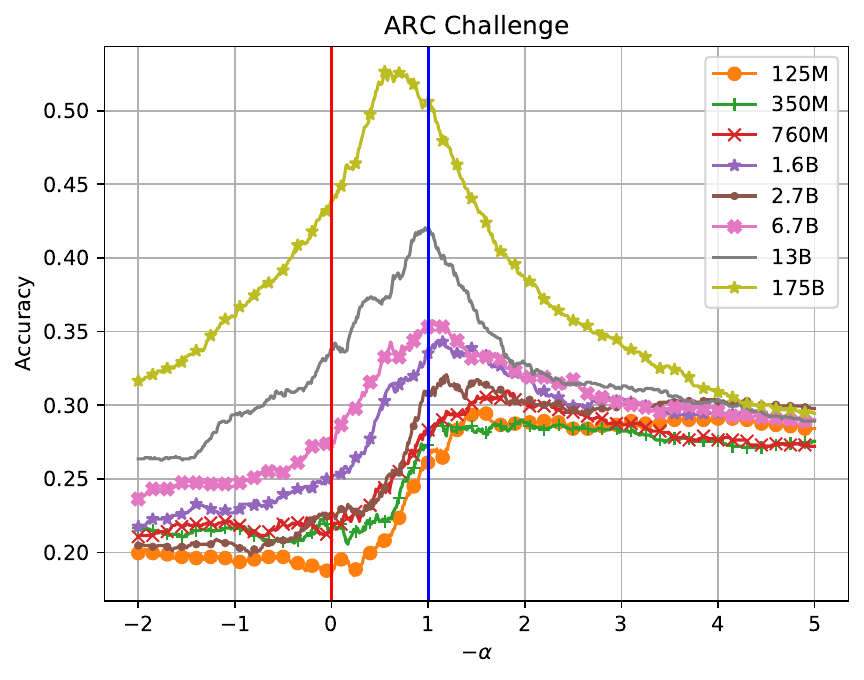}
    \end{subfigure}
    
        \vskip\baselineskip
    \begin{subfigure}[t]{.48\textwidth}
        \centering
        \includegraphics[width=.9\textwidth]{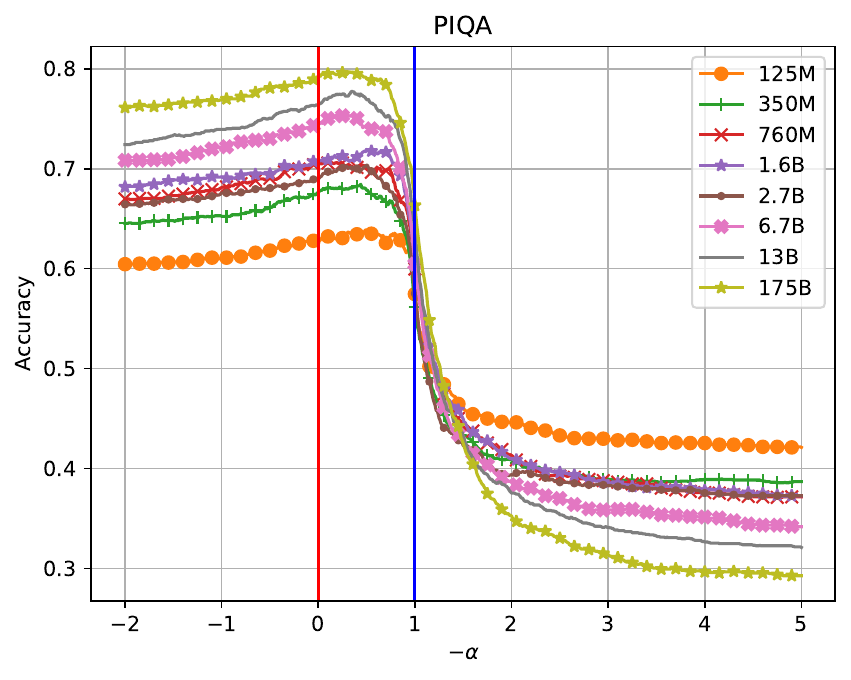}
    \end{subfigure}
    \hfill
    \begin{subfigure}[t]{.48\textwidth}
        \centering
        \includegraphics[width=.9\textwidth]{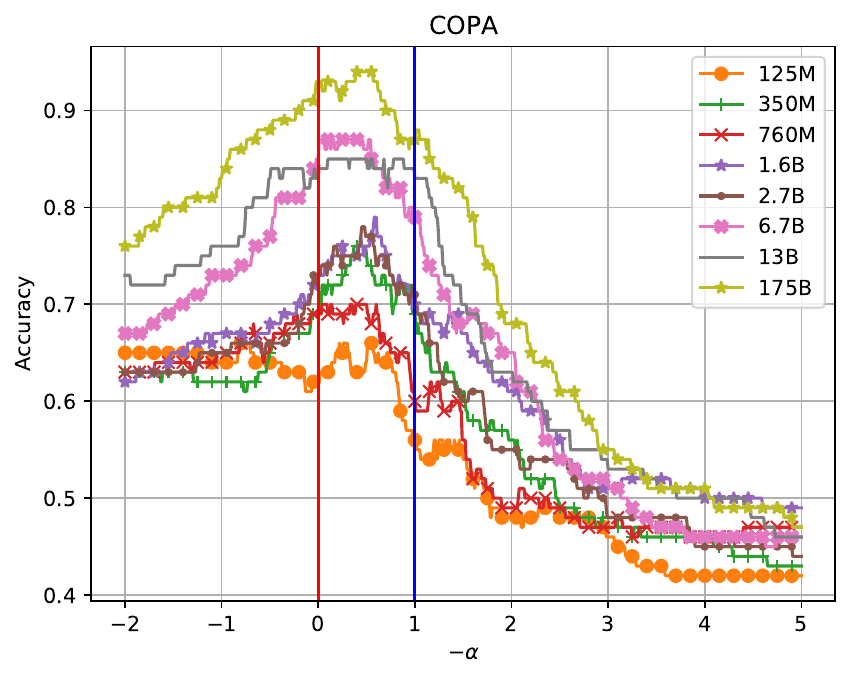}
    \end{subfigure}

\caption{Model comparison for StoryCloze, HellaSwag, OpenBookQA, CommonsenseQA, ARC Easy, ARC Challenge, PIQA and COPA by varying $\alpha$ on the testing set.}
\label{fig:multichoice_1}
\end{figure*}

\begin{figure*}[h]
\centering
    \begin{subfigure}[t]{.48\textwidth}
        \centering
        \includegraphics[width=.9\textwidth]{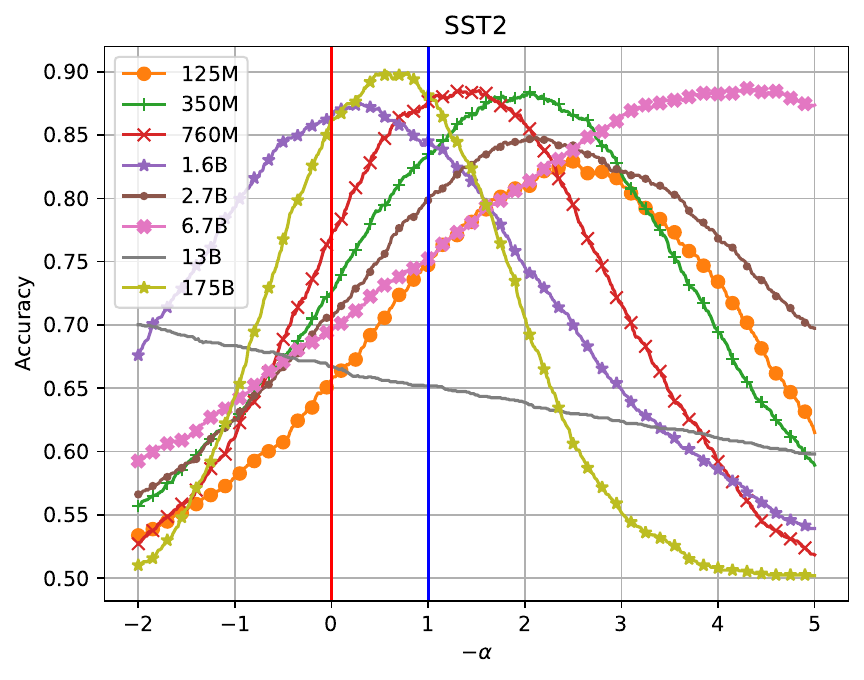}
    \end{subfigure}
    \hfill
    \begin{subfigure}[t]{.48\textwidth}
        \centering
        \includegraphics[width=.9\textwidth]{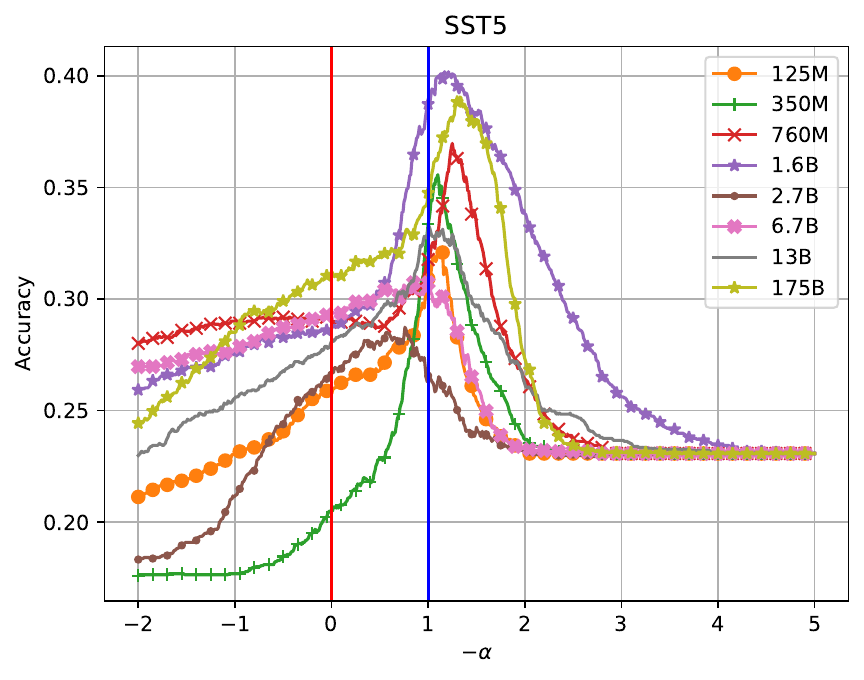}
    \end{subfigure}
    
    \vskip\baselineskip
    \begin{subfigure}[t]{.48\textwidth}
        \centering
        \includegraphics[width=.9\textwidth]{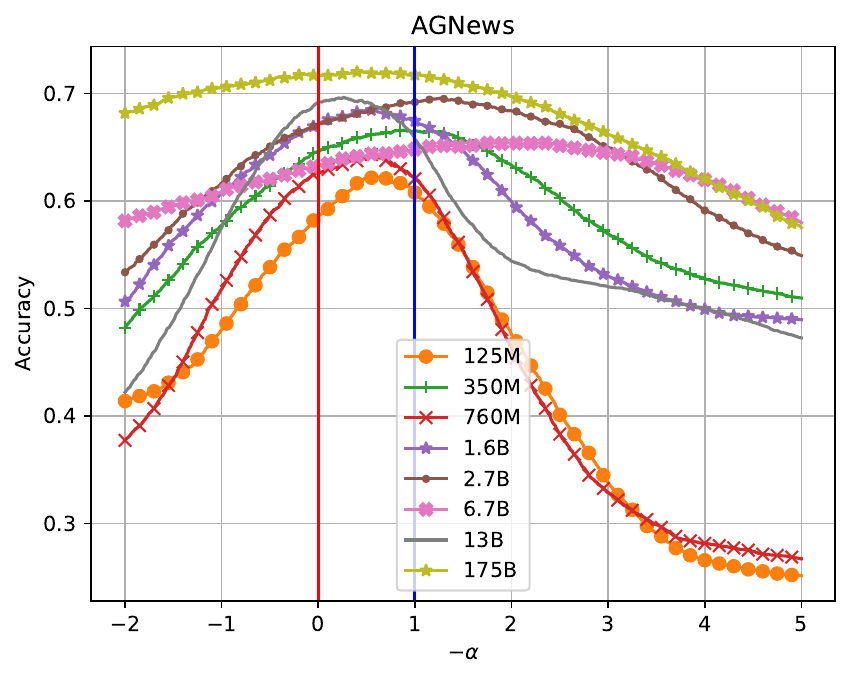}
    \end{subfigure}
    \hfill
    \begin{subfigure}[t]{.48\textwidth}
        \centering
        \includegraphics[width=.9\textwidth]{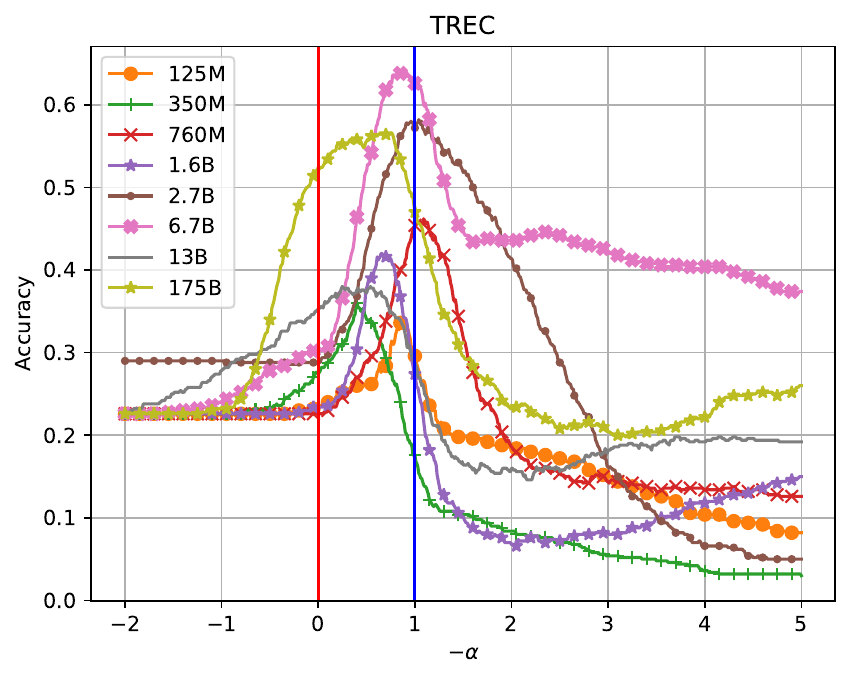}
    \end{subfigure}
    
        \vskip\baselineskip
    \begin{subfigure}[t]{.48\textwidth}
        \centering
        \includegraphics[width=.9\textwidth]{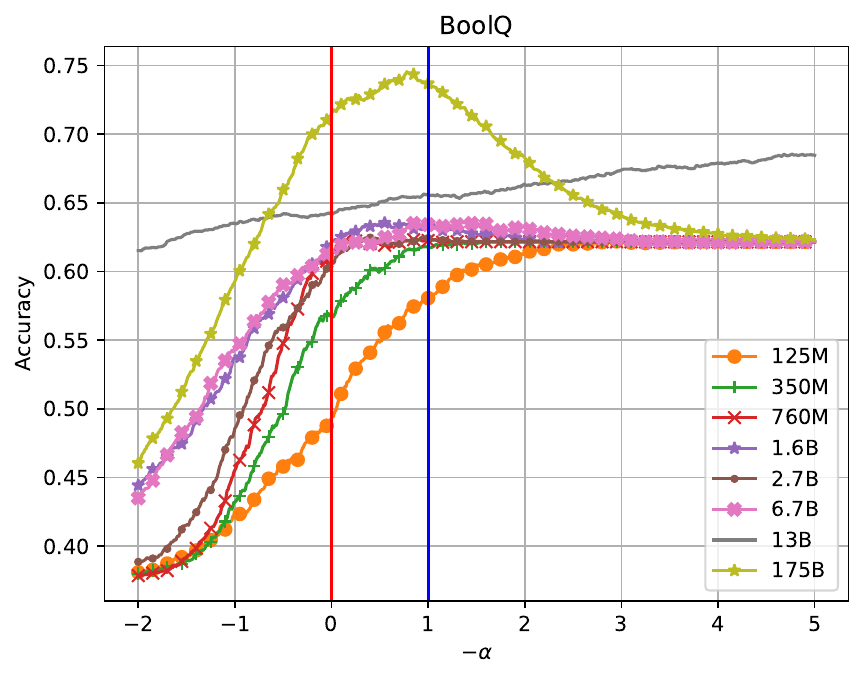}
    \end{subfigure}
    \hfill
    \begin{subfigure}[t]{.48\textwidth}
        \centering
        \includegraphics[width=.9\textwidth]{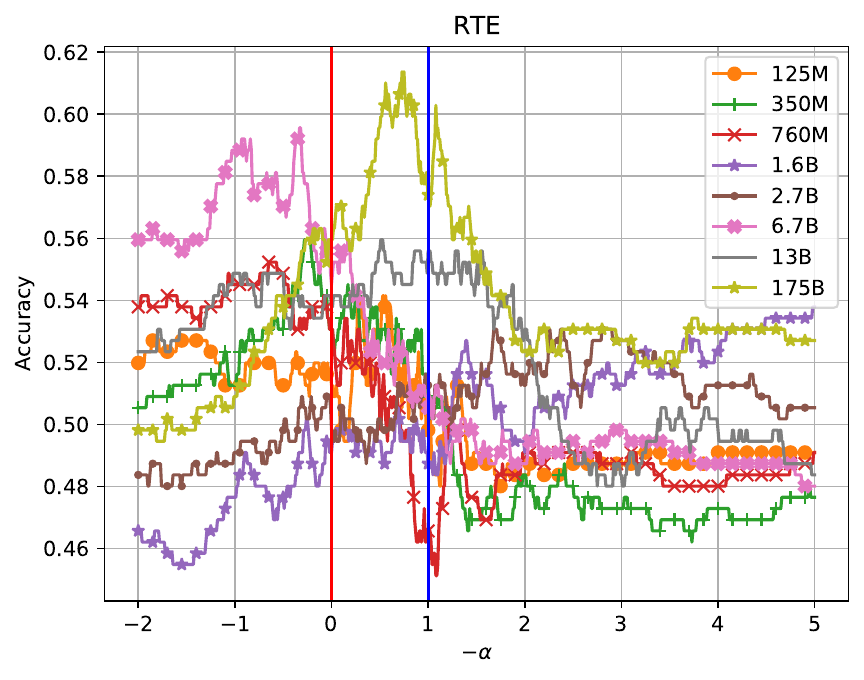}
    \end{subfigure}
    
    \vskip\baselineskip
    \begin{subfigure}[t]{.48\textwidth}
        \centering
        \includegraphics[width=.9\textwidth]{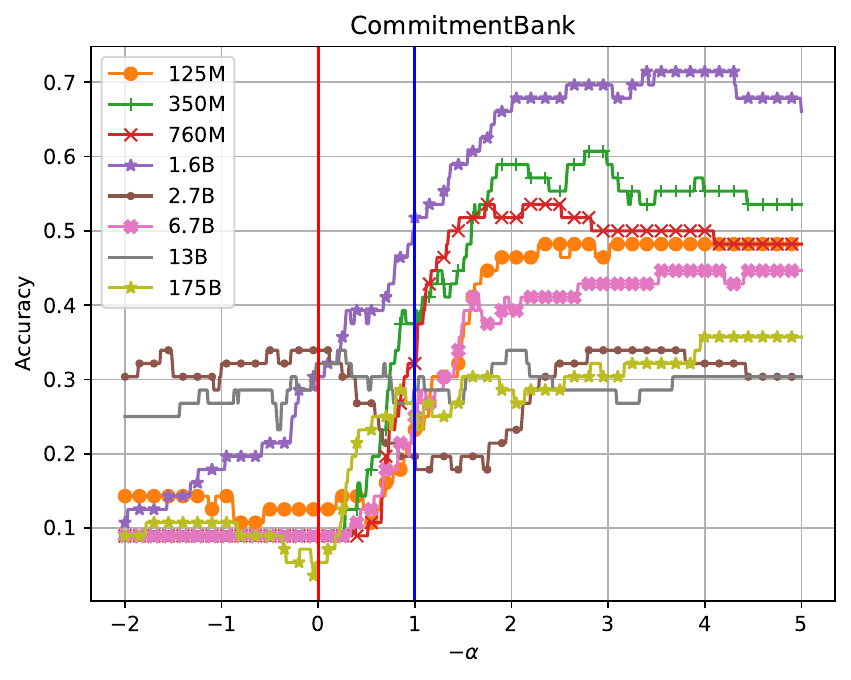}
    \end{subfigure}
\caption{Model comparison for SST-2, SST-5, AGNews, TREC, BoolQ, RTE and CommitmemtBank by varying $\alpha$ on the testing set.}
\label{fig:multichoice_2}
\end{figure*}

\end{document}